\PassOptionsToPackage{table}{xcolor}
\documentclass[sigconf,nonacm]{acmart}

\usepackage{xspace}
\usepackage{multirow}
\usepackage[capitalize]{cleveref}
\crefname{section}{Sec.}{Secs.}
\Crefname{section}{Section}{Sections}
\crefname{table}{Tab.}{Tabs.}
\Crefname{table}{Table}{Tables}
\crefname{appendix}{Sec.}{Secs.}
\Crefname{appendix}{Section}{Sections}
\crefname{subappendix}{Sec.}{Secs.}
\Crefname{subappendix}{Section}{Sections}

\definecolor{bestblue}{HTML}{CDE2FF}
\definecolor{secondblue}{HTML}{F1F7FE}
\newcommand{\bestcell}[1]{\cellcolor{bestblue}#1}
\newcommand{\secondcell}[1]{\cellcolor{secondblue}#1}

\newcommand{\eg}{e.g.\xspace}

\newcommand{\etal}{et~al.\xspace}

\newcommand{\method}{EvSync4D\xspace}

\DeclareMathOperator*{\argmin}{arg\,min}
\DeclareMathOperator*{\argmax}{arg\,max}
\DeclareMathOperator{\med}{med}
\DeclareMathOperator{\sign}{sign}
\DeclareMathOperator{\artanh}{artanh}

\newcommand{\Kc}{\mathbf{K}_{c}}          
\newcommand{\dc}{\mathbf{d}_{c}}          
\newcommand{\Ke}{\mathbf{K}_{e}}          
\newcommand{\de}{\mathbf{d}_{e}}          
\newcommand{\Rce}{\mathbf{R}}             
\newcommand{\tce}{\mathbf{t}}             
\newcommand{\Par}{\Theta}                 
\newcommand{\proje}{\pi_{e}}              
\newcommand{\projc}{\pi_{c}}              
\newcommand{\Fund}{\mathbf{F}}            
\newcommand{\Tc}{T_{c}}                   
\newcommand{\Te}{T_{e}}                   
\newcommand{\dt}{\Delta t}                
\newcommand{\dtpre}{\Delta t_{\mathrm{pre}}} 
\newcommand{\dtfin}{\Delta t^{\star}}     
\newcommand{\res}{\varepsilon}            
\newcommand{\lag}{\ell}                   
\newcommand{\twin}{\Delta_{\tau}}         
\newcommand{\dwin}{\Delta_{L}}            
\newcommand{\Wpol}{W}                     
\newcommand{\Tex}{T^{\mathrm{exp}}}       
\newcommand{\Trs}{T_{\mathrm{rs}}}        
\newcommand{\drift}{\xi}                  
\newcommand{\Score}{S}                    
\newcommand{\Ev}{\mathcal{E}}             
\newcommand{\Epol}{I^{E}}                
\newcommand{\Ecnt}{I^{E,\mathrm{cnt}}}    
\newcommand{\xc}{\mathbf{x}^{c}}          
\newcommand{\xe}{\mathbf{x}^{e}}          
\newcommand{\Corr}{\mathcal{C}}           
\newcommand{\Bd}{\mathbf{B}}              
\newcommand{\Hb}{\mathbf{H}}              
\newcommand{\Grid}{\mathcal{G}}           
\newcommand{\Lepi}{\mathcal{L}_{\mathrm{epi}}}
\newcommand{\Lcal}{\mathcal{L}_{\mathrm{cal}}}
\newcommand{\Lfd}{\mathcal{L}_{\mathrm{4D}}}
\newcommand{\Lsta}{\mathcal{L}_{\mathrm{sta}}}
\newcommand{\Lacc}{\mathcal{L}_{\mathrm{acc}}}
\newcommand{\Ldep}{\mathcal{L}_{\mathrm{dep}}}
\newcommand{\Lshp}{\mathcal{L}_{\mathrm{shp}}}
\newcommand{\Levt}{\mathcal{L}_{\mathrm{evt}}}
\newcommand{\Lsub}{\mathcal{L}_{\mathrm{sub}}}

\title{No Trigger, No Clock: \\ Calibrating Event--RGB Cameras for Dynamic 3D Geometry}

\author{Wanqi Yuan}
\affiliation{%
  \institution{Clemson University}
  \city{Clemson}
  \state{SC}
  \country{USA}}

\author{Brendan Desjardins}
\affiliation{%
  \institution{Clemson University}
  \city{Clemson}
  \state{SC}
  \country{USA}}

\author{Nianyi Li}
\affiliation{%
  \institution{Clemson University}
  \city{Clemson}
  \state{SC}
  \country{USA}}

\renewcommand{\shortauthors}{Yuan et al.}

\begin{document}

\begin{abstract}
An event camera at a second viewpoint can constrain the depth of moving objects, but exploiting independently recorded event and RGB streams is difficult: the temporal offset is unknown, the pair is uncalibrated, and events have no direct counterpart in RGB images.
We present \method, a framework that calibrates such a pair from an ordinary moving checkerboard, without a shared trigger or clock, and uses calibrated events to refine the metric geometry of moving objects.
We first recover the recording offset by correlating cross-modal motion and refine it to sub-frame precision jointly with the epipolar geometry.
Board poses from the RGB view then predict the moving grid, and aligning raw events with its projected lines jointly estimates the event-camera intrinsics and distortion, the metric extrinsics, and a residual offset, with no corner detection in the event view, intensity reconstruction, or task-specific training.
On the calibrated static rig, sparse event correspondences correct the scale and local errors of a monocular depth prior, while pretrained RGB tracks, event-view consistency, and local shape preservation regularize moving-object geometry over time.
Experiments on synthetic and real data demonstrate higher calibration accuracy and improved depth estimation accuracy for moving objects.

\end{abstract}

\maketitle

\section{Introduction}

\begin{figure}[t]
    \centering
    \includegraphics[width=0.8\columnwidth]{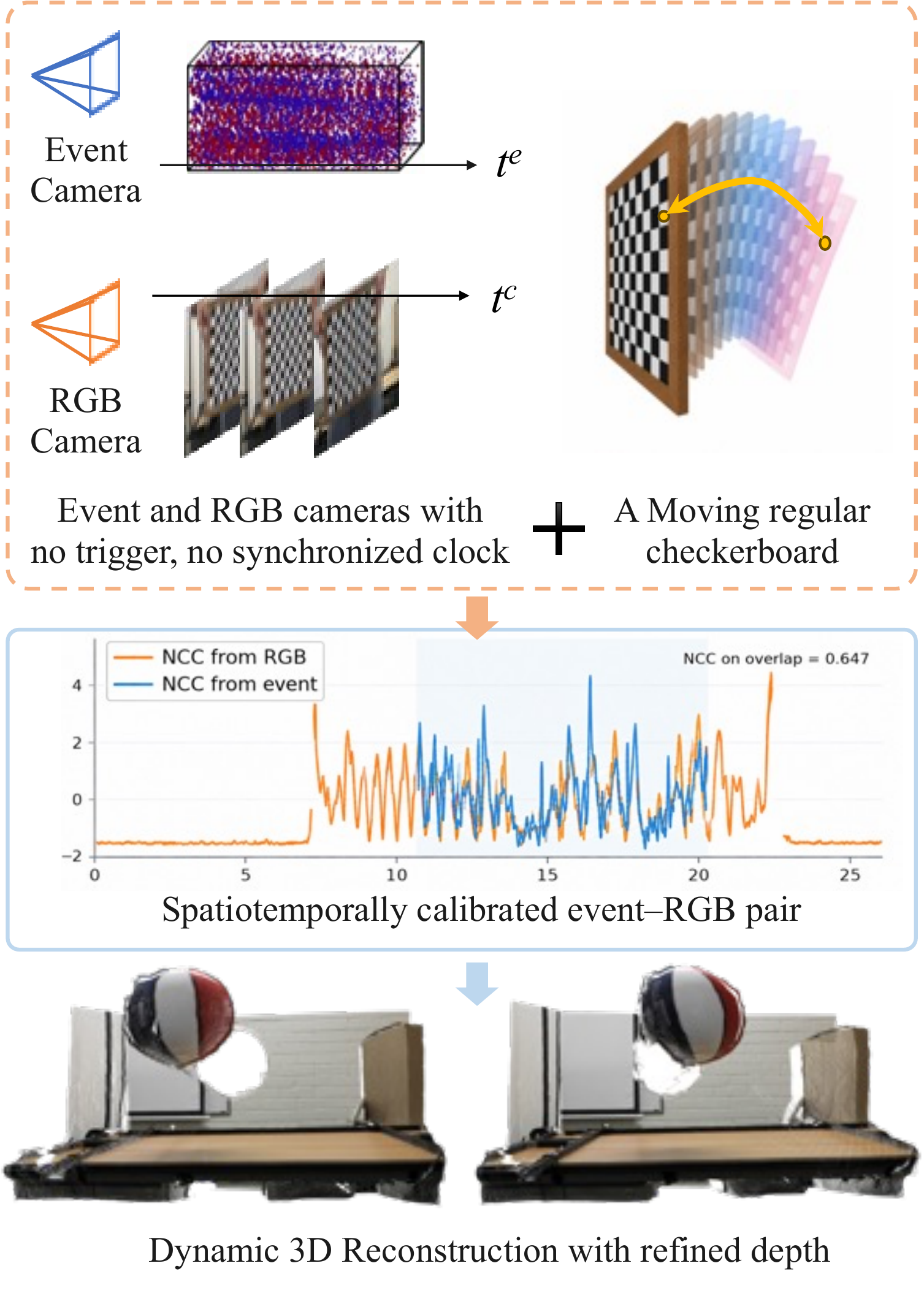}
    \vspace{-10pt}
    \caption{\textbf{\method} takes an event camera and an RGB camera that record without a shared trigger or clock, together with a moving checkerboard (top), calibrates the pair in time and space (middle), and uses the calibrated events to refine depth for dynamic 3D reconstruction (bottom).}
    \label{fig:teaser}
    \vspace{-10pt}
\end{figure}

Recovering the metric 3D geometry of moving objects from RGB video remains challenging.
Geometry foundation models such as Depth Anything~3~\cite{lin2511depth} and MoGe-3~\cite{kong2026moge} provide strong depth priors, but a single RGB viewpoint leaves ambiguity in metric scale and in the relation between object depth and motion.
Motion blur further weakens the image cues needed to recover local shape.
An event camera at a second viewpoint offers complementary measurements: cross-view correspondences constrain depth through the calibrated baseline, while asynchronous brightness changes with microsecond timestamps~\cite{gallego2020event} capture motion during and between RGB exposures.
Using these measurements to refine dense RGB geometry requires a common metric coordinate frame and time axis.

In practice, this prerequisite is mostly assumed rather than solved.
Event--RGB methods are typically developed on hybrid sensors, which deliver frames and events through the same low-resolution pixels (e.g., $346 \times 260$ for the DAVIS346~\cite{gallego2020event}), or on rigs synchronized and calibrated in advance~\cite{gehrig2021dsec, chang2026mertracker}, and take temporally aligned events as given~\cite{wu2025eag3r}.
Pairing an event camera with an ordinary high-resolution RGB camera provides neither shared pixels nor a shared clock, and poses two problems.
In time, the recordings start independently, so their timestamps differ by an unknown offset that can span many frames, whereas calibration methods assume synchronized data~\cite{muglikar2021calibrate} or refine the offset starting from zero~\cite{wang2024ef}.
In space, a moving target produces sparse, asynchronous events along its edges rather than an image with detectable corners, so existing methods depend on intensity reconstruction~\cite{muglikar2021calibrate}, active targets~\cite{wang2024motion, gossard2024ewand, hessenthaler2026simplified}, or dedicated patterns or corners detected in the event stream~\cite{huang2021dynamic, ryu2026corner2tag}.
The two problems are coupled: spatial calibration needs simultaneous observations of a target, whereas precise temporal alignment needs geometric constraints between views.

We present \method, which synchronizes and calibrates an event--RGB pair from two independently started recordings of an ordinary printed checkerboard, then uses calibrated events to refine the metric depth and local geometry of moving objects.
Our key observation is that \textbf{a moving checkerboard addresses both problems}: it is a motion signal shared by the two modalities and a geometric reference for raw events.
Our contributions are:
\begin{itemize}
    \item \textbf{Coarse-to-fine temporal alignment.}
    As a motion signal, the board yields the recording offset before any correspondence exists: we correlate the image motion tracked in the RGB video and in event frames, then match and track points across the modalities and refine the offset to sub-frame precision jointly with the epipolar geometry.
    \item \textbf{RGB-anchored raw-event calibration.}
    As a geometric reference, the board needs no detection in the event view: its poses in the RGB frames predict where each grid line projects in the event camera at any instant, i.e., where the moving board triggers events, and we fit every raw event to the predicted line at its own timestamp.
    This jointly estimates the event-camera intrinsics and distortion, the metric extrinsics, and a residual offset, with no intensity reconstruction, active target, or task-specific training.
    \item \textbf{Event-constrained dynamic 3D geometry.}
    On a static rig with one RGB camera and an event camera at a second viewpoint, we use sparse event correspondences to correct the scale and local errors of a dense monocular depth prior.
    We refine moving-object geometry over time using pretrained RGB tracks, event-view consistency, and local shape preservation, without task-specific training.
    As a downstream application, the refined geometry and event-view tracks support novel-view slow-motion visualization between RGB frames.
\end{itemize}

\section{Related Work}
\noindent \textbf{Event Camera Calibration.}
Checkerboard calibration~\cite{zhang2000flexible} does not transfer to event cameras, where a static target produces no events and a moving one no image~\cite{sundar2024generalized}.
Prior methods use blinking LEDs or flashing screens as targets~\cite{cai2024accurate, wang2024motion, gossard2024ewand, hessenthaler2026simplified}, which are well localized but require an active device; calibrate on images reconstructed by a learned network~\cite{rebecq2019high, muglikar2021calibrate}, whose artifacts propagate into the detected corners~\cite{zhang2022formulating}; or detect printed patterns in the event stream: circle grids~\cite{huang2021dynamic, salah2024calib, chen2025ekalibr}, whose centers are biased under lens distortion~\cite{song2025camera}, and recently checkerboard corners and tags in images of warped events~\cite{ryu2026corner2tag, gallego2018unifying}.
All of them localize target features in the event view; we detect the checkerboard only in the RGB camera and fit raw events to the board edges predicted by its pose.

\vspace{3pt} 
\noindent \textbf{Spatiotemporal Calibration.}
Continuous-time trajectories make the temporal offset between sensors a differentiable parameter~\cite{furgale2013unified, rehder2016general, sommer2020efficient} and suit asynchronous events~\cite{mueggler2018continuous}.
EF-Calib~\cite{wang2024ef} jointly calibrates the intrinsics, extrinsics, and temporal offset of an event--frame pair, and eKalibr-Stereo~\cite{chen2025ekalibrstereo} extends continuous-time bundle adjustment to event-based stereo rigs; both rely on target features detected in the event view.
Without a target, unsynchronized videos can be aligned by matching point trajectories across cameras~\cite{caspi2002spatio, padua2008linear}.
We extend this idea across modalities with modern point trackers and matchers~\cite{karaev2025cotracker3, lindenberger2023lightglue}, which recover offsets of many frames before any spatial calibration, and refine the remaining sub-frame offset with raw events.
 
\vspace{3pt} 
\noindent \textbf{Event-Based Dynamic 3D Geometry.}
Events, long used for deblurring~\cite{pan2019bringing} and frame interpolation~\cite{tulyakov2021time, ma2024timelens}, now also serve dynamic 3D reconstruction: EAG3R~\cite{wu2025eag3r} and Weerakoon~\etal~\cite{weerakoon2026events} augment feed-forward geometry models with events, Interp3R~\cite{guo2026interp3r} interpolates geometry between frames, ERF-GS~\cite{bai2026erfgs} optimizes dynamic Gaussians from events and frames with disjoint viewpoints, ARES~\cite{lee2026ares} estimates metric scene flow from asymmetric event--RGB stereo, and MER-Tracker~\cite{chang2026mertracker} learns high-speed 3D point tracking from a synchronized multi-view rig.
We calibrate a single, independently recorded event--RGB pair and use its events as sparse metric constraints on moving-object depth and local geometry, with pretrained tracks providing temporal correspondences.

\section{Method}
\label{sec:method}

\subsection{Problem Setup and Overview}
\label{sec:overview}

\noindent \textbf{Setup.}
A rigid, static rig holds an RGB camera and an event camera that record independently, without a shared trigger or clock.
The RGB camera has known intrinsics $\Kc$ and distortion $\dc$~\cite{zhang2000flexible} and delivers frames $\{I_k\}$ with period $\Tc$ at the times $t_k = k\,\Tc$ on its own clock.
The event camera delivers a stream $\Ev$ of events $(\mathbf{x}_n, t_n, p_n)$ with pixel location $\mathbf{x}_n$, timestamp $t_n$ on its own clock, and polarity $p_n \in \{-1,+1\}$.
We relate the two readings $t^{c}$ and $t^{e}$ of the same instant by a constant, unknown recording offset $\dt$,
\begin{equation}
    t^{c} = t^{e} + \dt ,
    \label{eq:temporal_offset}
\end{equation}
which neglects clock drift and the exposure and rolling shutter of the RGB camera; their effect is bounded for recordings of a few seconds.
Besides $\dt$, we estimate the intrinsics $\Ke$ and distortion $\de = (k_1, k_2, p_1, p_2)$ of the event camera and the rigid transformation $\mathbf{X}^{e} = \Rce\,\mathbf{X}^{c} + \tce$ from RGB- to event-camera coordinates, using only an ordinary printed checkerboard with square size $a$, moved by hand.

\begin{figure*}[t]
    \centering
    \includegraphics[width=\linewidth]{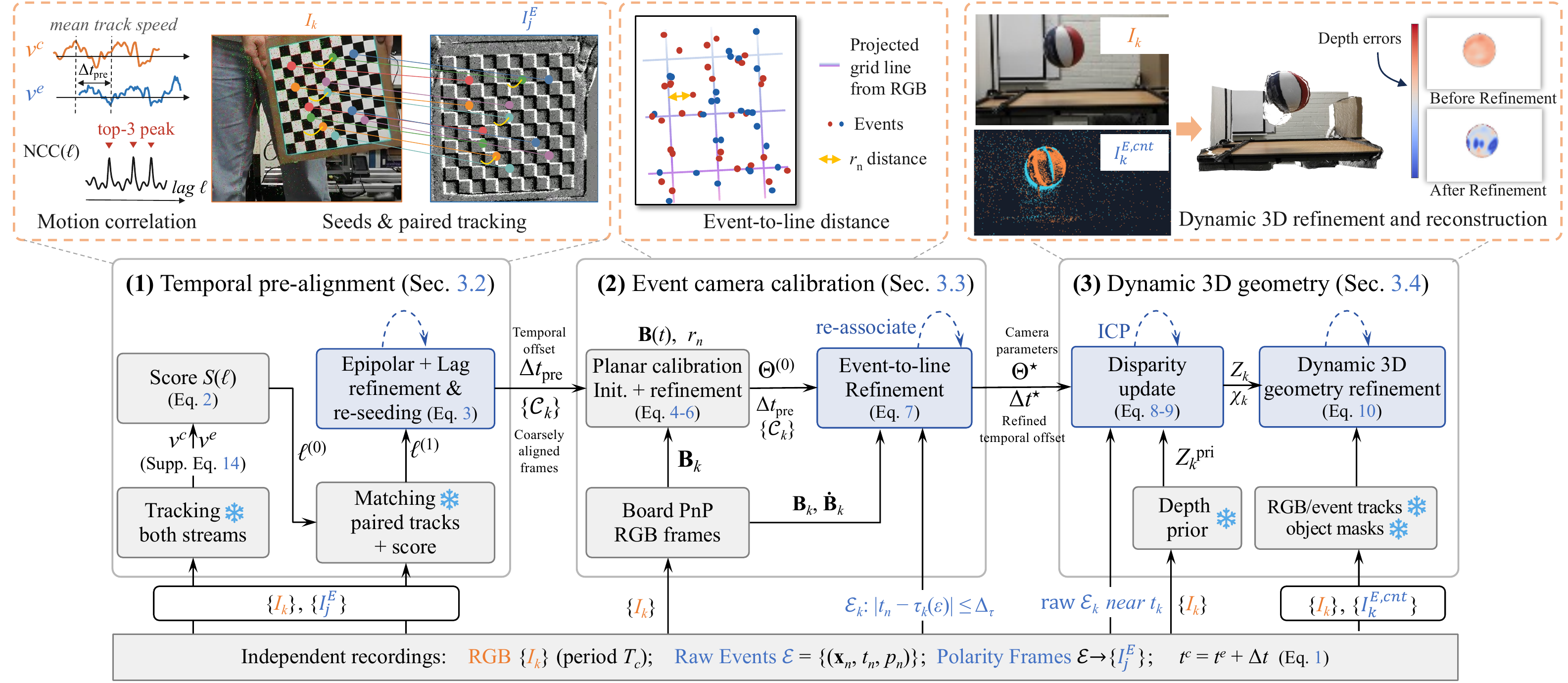}
    \vspace{-15pt}
    \caption{\textbf{Overview of \method.}
    \emph{Temporal pre-alignment} (\cref{sec:prealign}) aligns the two independent recordings up to a fraction of a frame.
    \emph{Raw-event calibration} (\cref{sec:calib}) registers raw events to the grid lines predicted from the board poses in the RGB frames and estimates $\Ke$, $\de$, $\Rce$, $\tce$, and the final offset $\dtfin$.
    The calibrated rig supports event-guided metric depth and temporal geometry refinement, with sub-frame visualization as a downstream application (\cref{sec:fourd}).
    Snowflakes mark means frozen pretrained networks.}
    \label{fig:pipeline}
    \vspace{-15pt}
\end{figure*}

\vspace{3pt}
\noindent \textbf{Overview.}
\method proceeds from coarse to fine in time and uses the events at increasing temporal granularity (\cref{fig:pipeline}).
\textbf{(1)~Temporal pre-alignment} (\cref{sec:prealign}) recovers $\dt$ up to a fraction of a frame from motion that is visible in both views, together with sparse cross-modal correspondences.
\textbf{(2)~RGB-anchored raw-event calibration} (\cref{sec:calib}) registers the raw events to the grid lines of the moving board, which are predicted from the RGB frames, and estimates $\Ke$, $\de$, $\Rce$, $\tce$, and a residual offset $\res$.
\textbf{(3)~Event-constrained dynamic 3D geometry} (\cref{sec:fourd}) uses the calibrated rig to correct a monocular depth prior and refine moving-object geometry along pretrained tracks; sub-frame visualization is a downstream application.
All learned components are pretrained and frozen.
Before calibration, \emph{polarity frames} $\Epol_j$ encode each pixel's last-event polarity for tracking, matching, and re-seeding; $j$ indexes event frames.
After calibration, \emph{count frames} $\Ecnt_k$ count events per pixel around RGB time $t_k$.
\Cref{sec:supp_setup} details both renderings and the error of \cref{eq:temporal_offset}; \cref{sec:supp_diagram} adds a block diagram.

\subsection{Motion-Based Temporal Pre-Alignment}
\label{sec:prealign}

The offset can span many frames, so correspondences cannot be trusted before a rough alignment.
We first align a correspondence-free motion signal, then establish correspondences, and finally refine the offset to sub-frame precision jointly with the epipolar geometry (\cref{fig:pipeline}, top left).

\vspace{3pt}
\noindent \textbf{Coarse lag from motion correlation.}
CoTracker3~\cite{karaev2025cotracker3} independently tracks RGB frames $\{I_k\}$ from checkerboard corners and polarity frames $\{\Epol_j\}$ from a regular grid; these points need not correspond.
Mean displacements of visible tracks yield motion signals $v^{c}[k]$ and $v^{e}[k]$ on the RGB period $\Tc$, after resampling event tracks.
For an integer lag $\lag$, i.e., $\dt = \lag\,\Tc$, let $\mathrm{NCC}(\lag)$ be the normalized cross-correlation of $v^{c}[k+\lag]$ and $v^{e}[k]$ over their $N_{\lag}$ common samples.
Since a short overlap can produce a high correlation by chance, we score a lag by the Fisher $z$-statistic~\cite{fisher1921probable},
\begin{equation}
    \Score(\lag) = \artanh\!\big(\mathrm{NCC}(\lag)\big)\, \sqrt{N_{\lag} - 3} ,
    \label{eq:score}
\end{equation}
which measures the significance of a correlation rather than its size, and take $\lag^{(0)} = \argmax_{\lag} \Score(\lag)$.
A board waved back and forth produces side peaks of $\Score$ at multiples of its period, so the next step validates the lag geometrically.

\vspace{3pt}
\noindent \textbf{Cross-modal matching and paired tracks.}
Given $\lag^{(0)}$, we pair $\Epol_j$ with the RGB frame nearest its time shifted by $\lag^{(0)}\Tc$.
We render the RGB frame such that it resembles a polarity frame: as a crop around the detected board and as an event-like image of the signed log-intensity difference between its two neighboring frames.
We match keypoints with SuperPoint~\cite{detone2018superpoint} and LightGlue~\cite{lindenberger2023lightglue} and filter the matches with a RANSAC homography, as the board is planar.
Frame pairs with the most and best-spread inliers seed CoTracker3 queries in $\{I_k\}$ and the same $\{\Epol_j\}$, yielding \emph{paired tracks} $\{(\xc_i[k], \xe_i[k])\}_{i=1}^{M}$ whose common index denotes the same scene point.
Evaluating $\Score$ on them, where both signals now measure the same points, gives the lag $\lag^{(1)}$, which we accept if a fundamental matrix fitted at this lag has enough RANSAC inliers: at a lag off by one period of the motion, the tracks of a pair show the board in different poses and violate the epipolar constraint.

\vspace{3pt}
\noindent \textbf{Joint epipolar--temporal refinement.}
Because the rig is static, simultaneous observations of any scene point satisfy $\tilde{\mathbf{x}}^{c\top} \Fund\, \tilde{\mathbf{x}}^{e} = 0$, in homogeneous coordinates, for a single fundamental matrix $\Fund$~\cite{hartley2003multiple}.
For a moving point, an error in the lag displaces the RGB observation along its image trajectory and violates this constraint, so the epipolar residual is a continuous function of the lag.
We relax $\lag$ to a real number, interpolate the RGB tracks linearly, and minimize
\begin{equation}
    \Lepi(\Fund, \lag) = \frac{1}{|\Omega_{\lag}|} \sum_{(i,k) \in \Omega_{\lag}} d_{\mathrm{S}}\big(\xc_i(k+\lag),\, \xe_i[k];\, \Fund\big)
    \label{eq:joint_refine}
\end{equation}
over $\Fund$ and $\lag$, where $d_{\mathrm{S}}$ is the squared Sampson distance~\cite{hartley2003multiple} and $\Omega_{\lag}$ contains the track--sample pairs $(i,k)$ visible in both views at lag $\lag$; a hinge penalty keeps this overlap from shrinking.
We initialize $\Fund$ with RANSAC at $\lag^{(1)}$ and optimize with Adam; the minimizer $\lag^{(2)} \in \mathbb{R}$ gives the pre-aligned offset $\dtpre = \lag^{(2)}\,\Tc$.
As all frames share the same $\Fund$, the planar board is not degenerate if it changes depth and orientation; it should move in several directions, since only motion across epipolar lines constrains the lag.

\vspace{3pt}
\noindent \textbf{Per-frame correspondences.}
Paired tracks end when points become occluded or leave the field of view.
When few pairs remain, we re-match $I_k$ with the closest $\Epol_j$ under $\dtpre$, retain epipolar inliers, and re-track both sequences, yielding for every RGB frame $k$ the cross-modal pairs $\Corr_k = \{(\xc_{k,h}, \xe_{k,h})\}_h$: pair $h$ links a point in $I_k$ to the same scene point in the closest polarity frame.
\Cref{sec:supp_prealign} details all steps and shows a real example.

\subsection{RGB-Anchored Raw-Event Calibration}
\label{sec:calib}

Event frames blur time over their accumulation window, and $\Fund$, which is estimated without a distortion model, is projective rather than metric.
We therefore return to raw events and let the RGB camera, where the checkerboard is easy to detect, anchor the geometry: the board poses in the RGB frames predict where every grid line projects in the event camera at any instant, and the events of the moving board must lie on these lines at their own timestamps (\cref{fig:pipeline}, top middle).
Calibration thus becomes a registration of raw events to predicted lines over $\Par = \{\Ke, \de, \Rce, \tce, \res\}$ and needs no corner detection in the event view.

\vspace{3pt}
\noindent \textbf{Continuous-time board model.}
For each RGB frame $k$ with detected corners, planar PnP gives the first two columns $\mathbf{r}^{1}_k, \mathbf{r}^{2}_k$ of the board rotation and the board origin $\mathbf{o}_k$ in RGB-camera coordinates.
A board point with the in-plane coordinates $\mathbf{u} = (u, v)^{\top}$, homogeneously $\tilde{\mathbf{u}} = (u, v, 1)^{\top}$, then lies at $\mathbf{X}^{c} = \Bd_k \tilde{\mathbf{u}}$ with $\Bd_k = [\,\mathbf{r}^{1}_k \;\; \mathbf{r}^{2}_k \;\; \mathbf{o}_k\,]$, and central differences give the board velocity $\dot{\Bd}_k$.
On the event clock, frame $k$ was captured at $\tau_k(\res) = t_k - \dtpre + \res$, where the residual offset $\res$ absorbs the sub-frame error left in $\dtpre$.
Around it, we use the first-order motion model
\begin{equation}
    \Bd(t) = \Bd_k + \big(t - \tau_k(\res)\big)\, \dot{\Bd}_k , \qquad |t - \tau_k(\res)| \leq \twin ,
    \label{eq:board_motion}
\end{equation}
with a half-width $\twin$ of a few milliseconds.
A board point $\mathbf{u}$ observed at the event time $t$ then has the event-camera coordinates $\Hb(t)\,\tilde{\mathbf{u}}$ and the predicted pixel location
\begin{equation}
    \hat{\mathbf{x}}(\mathbf{u}, t; \Par) = \proje\big(\Hb(t)\,\tilde{\mathbf{u}}\big), \;\;\; \Hb(t) = \Rce\,\Bd(t) + \tce\,\mathbf{e}_3^{\top} ,
    \label{eq:event_projection}
\end{equation}
where $\mathbf{e}_3 = (0,0,1)^{\top}$ and $\proje$ is the pinhole projection with intrinsics $\Ke$ and radial--tangential distortion $\de$.
Changing $\res$ slides every predicted line along its image velocity, so $\res$ is observable whenever the board moves.

\vspace{3pt}
\noindent \textbf{Coarse initialization.}
Registering events to a periodic grid requires an initial guess within half a square.
We obtain it from $\Corr_k$: the viewing ray of an RGB point $\xc_{k,h}$ intersects the board plane given by $\Bd_k$ at the board coordinates $\mathbf{u}_{k,h}$, so the pairs $(\mathbf{u}_{k,h}, \xe_{k,h})$ act as planar 3D--2D correspondences for the event camera.
Zhang's closed-form solution~\cite{zhang2000flexible} then gives an initial $\Ke$ and the board pose in the event camera for every frame; composing these poses with the board poses in the RGB camera gives $\Rce$ and $\tce$, i.e., the initial $\Par^{(0)}$.

\vspace{3pt}
\noindent \textbf{Event-to-line residual.}
For each RGB frame $k$, we select the raw events $\mathcal{E}_k = \{n : |t_n - \tau_k(\res)| \leq \twin\}$ inside the detected board region, which excludes the events of the hand that holds the board.
A moving checkerboard triggers events along its interior grid lines $\Grid$, where black and white squares meet.
We back-project each event onto the board plane at its own timestamp, $\tilde{\mathbf{u}}_n \sim \Hb(t_n)^{-1}\, \proje^{-1}(\mathbf{x}_n)$, and associate it with the nearest grid line $g_n \in \Grid$, on which $\mathbf{q}_n$ is the foot point of $\mathbf{u}_n$.
The residual of event $n$ is its signed distance, in the event image, to the projected line,
\begin{equation}
    r_n(\Par) = \mathbf{n}_n^{\top} \big( \mathbf{x}_n - \hat{\mathbf{x}}(\mathbf{q}_n, t_n; \Par) \big) ,
    \label{eq:line_distance}
\end{equation}
where $\mathbf{n}_n$ is the unit normal of the projection of $g_n$ at $\hat{\mathbf{x}}(\mathbf{q}_n, t_n; \Par)$.
Each event thus contributes one scalar constraint at its own timestamp, without an accumulation window and regardless of its polarity, since the ON and OFF events of an edge lie on the same line.
As in ICP~\cite{besl1992method}, we alternate between associating the events and optimizing $\Par$, and reject events with large residuals.

\vspace{3pt}
\noindent \textbf{Spatiotemporal refinement.}
We estimate $\Par$ in three stages $s = 1, 2, 3$ of increasing complexity, each minimizing
\begin{equation}
\begin{split}
    & \Lcal^{(s)}(\Par) = \sum_{k} \sum_{n \in \mathcal{E}_k} \rho_{\sigma_s}\big( r_n(\Par) \big) \\
    & \quad + \beta_s \sum_{k} \sum_{h \in \Corr_k} \rho_{\sigma_s}\big( \| \hat{\mathbf{x}}(\mathbf{u}_{k,h}, \tau_k; \Par) - \xe_{k,h} \| \big) ,
\end{split}
\label{eq:calib_objective}
\end{equation}
where $\rho_{\sigma}(r) = \sigma^{2} \log(1 + r^{2}/\sigma^{2})$ is the Cauchy loss, whose scale $\sigma_s$ decreases over the stages.
The first term registers the raw events to the predicted grid lines; the second reuses the sparse correspondences, which are less accurate but not periodic and thus keep the solution from shifting by an entire square while the registration is still coarse.
The first stage optimizes $\Ke$, $\Rce$, and $\tce$ with $\beta_1 > 0$; the second adds the radial coefficients $(k_1, k_2)$, and the third the tangential coefficients $(p_1, p_2)$ and $\res$, both with $\beta_s = 0$.
Since the square size $a$ is known, $\tce$ is metric, and the final recording offset is $\dtfin = \dtpre - \res^{\star}$, with the final estimate $\res^{\star}$.
\Cref{sec:supp_calib_details} details all steps.

\subsection{Event-Constrained Dynamic 3D Geometry}
\label{sec:fourd}

With events aligned to the RGB clock by $\dtfin$, the calibrated rig supplies sparse metric constraints to correct a dense depth prior and refine moving-object geometry over time.
Pretrained tracks provide temporal correspondences for this refinement, and we can use the refined geometry for sub-frame visualization.

\vspace{3pt}
\noindent \textbf{Event-guided metric depth.}
Rectification relates RGB depth $Z$ to disparity $d = f b / Z$, with focal length $f$ and baseline $b = \|\tce\|$.
We refine the depth priors from 3D foundation models (\eg DA3 \cite{lin2511depth})
$Z^{\mathrm{pri}}_k$ through $d^{\mathrm{pri}}_k = f b / Z^{\mathrm{pri}}_k$.
From frame $k$'s log intensity $L_k$ and flow $\mathbf{w}_k$ to the next frame~\cite{teed2020raft}, the predicted change $\Delta L_k \approx -(2\dwin/\Tc)\, \nabla L_k^{\top} \mathbf{w}_k$ over a centered window of half-width $\dwin$ identifies event edges $\mathcal{P}_k$ (large $|\Delta L_k|$) and polarity $\sign \Delta L_k$~\cite{gallego2020event}.
Same-row, same-polarity event--edge pairs vote for a disparity scale $\alpha_m$ per prior inference chunk $m$, selected for consistency across chunks.
Nearest-edge matches under this scaled prior yield disparity $d_q$ at pixel $\mathbf{x}_q$ and log error $c_q = \log\!\big(d_q / (\alpha_m d^{\mathrm{pri}}_k(\mathbf{x}_q))\big)$ for match $q$.
We propagate these corrections to RGB pixels $\mathbf{x}$:
\begin{align}
    d_k(\mathbf{x}) &= \alpha_m\, d^{\mathrm{pri}}_k(\mathbf{x})\, \exp\!\left( \frac{\sum_{q} \kappa_q(\mathbf{x})\, c_q}{w_0 + \sum_{q} \kappa_q(\mathbf{x})} \right) ,
    \label{eq:depth_update} \\
    \chi_k(\mathbf{x}) &= \frac{\sum_{q} \kappa_q(\mathbf{x})}{w_0 + \sum_{q} \kappa_q(\mathbf{x})} \;\in\; [0, 1) .
    \label{eq:confidence}
\end{align}
Here $\kappa_q(\mathbf{x})$ weights spatial and prior-depth proximity, the prior weight $w_0$ retains the scaled prior away from matches, and event confidence $\chi_k(\mathbf{x})$ is the relative event weight.
Metric depth is $Z_k = f b / d_k$; blurred surfaces select disparity by event-based deblurring~\cite{pan2019bringing}, and static background uses temporal-median disparity (\cref{sec:supp_depth}).

\vspace{3pt}
\noindent \textbf{Temporal geometry refinement.}
CoTracker3 tracks a dense RGB grid, giving image tracks $\mathbf{x}_i[k]$ lifted as $\mathbf{X}^{0}_i[k] = Z_k(\mathbf{x}_i[k])\, \projc^{-1}(\mathbf{x}_i[k])$, where $\projc^{-1}$ back-projects to a unit-depth RGB ray.
Following Stereo4D~\cite{jin2025stereo4d}, we optimize offsets $\zeta_i[k]$ along unit bearings $\mathbf{b}_i[k]$, $\mathbf{X}_i[k] = \mathbf{X}^{0}_i[k] + \zeta_i[k]\, \mathbf{b}_i[k]$, by minimizing
\begin{equation}
\begin{split}
    \Lfd = {} & \sum_{i} \big[ \mu_i\, \Lsta^{i} + (1 - \mu_i)\, \Lacc^{i} \big] + \lambda_{\mathrm{dep}}\, \Ldep \\
    & + \lambda_{\mathrm{shp}}\, \Lshp + \lambda_{\mathrm{evt}}\, \Levt .
\end{split}
\label{eq:fourd_objective}
\end{equation}
Stereo4D's $\Lsta^{i}$ and $\Lacc^{i}$ penalize positional spread and ray acceleration, balanced by the static-track indicator $\mu_i \in [0,1]$; $\lambda_{\mathrm{dep}}, \lambda_{\mathrm{shp}}, \lambda_{\mathrm{evt}}$ weight our three additions.
The \emph{source-aware depth term} $\Ldep$ anchors inverse ranges to those of $\mathbf{X}^{0}_i[k]$, weighted by $\omega_{\mathrm{pri}}^{\,1-\chi}\, \omega_{\mathrm{evt}}^{\,\chi}$, $\chi = \chi_k(\mathbf{x}_i[k])$, on moving surfaces and $\omega_{\mathrm{bg}}$ on static background, with $\omega_{\mathrm{evt}} > \omega_{\mathrm{bg}} > \omega_{\mathrm{pri}}$.
The \emph{shape term} $\Lshp$ applies a Cauchy loss to deviations of $D_{ij}[k] = \|\mathbf{X}_i[k] - \mathbf{X}_j[k]\|$ from its temporal median, over track pairs $(i,j) \in \mathcal{N}$ on the same SAM~3-segmented object~\cite{carion2026sam}.
The \emph{event-view term} $\Levt$ penalizes temporal variation of each moving point's distance to its partner's event-camera ray; partners are initialized by projection and tracked through $\{\Ecnt_k\}$ with CoTracker3.
Allowing nonzero distances accommodates imperfect initialization (\cref{sec:supp_traj}).

\vspace{3pt}
\noindent \textbf{Application: sub-frame visualization.}
Between consecutive RGB frames, event-view tracks at sub-frame times fix each moving point's event-camera ray (supplement video),
while the refined geometry at both frames and local shape preservation regularize its range along the ray.
Rendering these sub-frame trajectories yields novel-view 4D slow motion (\cref{sec:supp_subframe}).

\section{Experiments}
\label{sec:experiments}

Further details, the complete result tables, and additional results, including dynamic 3D reconstructions and 4D slow motion, can be found in the appendix.

\subsection{Experimental Setup}
\label{sec:exp_setup}

\begin{table*}[t]
\centering
\caption{\textbf{Event-camera intrinsics on synthetic data}. Relative focal-length error $e_f$ (\%), principal-point error $e_c$ (px), and projection error $e_{\pi}$ (px) with respect to the ground truth. One factor is varied at a time around the default configuration (20\,cm baseline, parallel cameras, no temporal offset), which is the 20\,cm row. Dark and light blue mark the best and second-best results for each metric within each case and each summary row, respectively. Summary statistics include all 9 cases. The estimated parameters are listed in \cref{tab:calib_syn_full}. $^\dagger$Our reimplementation, as the official code is not yet available.}
\label{tab:calib_syn}
\vspace{-5pt}
\footnotesize
\setlength{\tabcolsep}{6pt}
\begin{tabular}{ll ccc ccc ccc}
\toprule
& & \multicolumn{3}{c}{E2Calib~\cite{muglikar2021calibrate}}
& \multicolumn{3}{c}{Corner2Tag$^\dagger$~\cite{ryu2026corner2tag}}
& \multicolumn{3}{c}{\method (ours)} \\
\cmidrule(lr){3-5} \cmidrule(lr){6-8} \cmidrule(lr){9-11}
Factor & Value
& $e_f\downarrow$ & $e_c\downarrow$ & $e_{\pi}\downarrow$
& $e_f\downarrow$ & $e_c\downarrow$ & $e_{\pi}\downarrow$
& $e_f\downarrow$ & $e_c\downarrow$ & $e_{\pi}\downarrow$ \\
\midrule
\multirow{3}{*}{Temporal offset $\dt$}
& $-0.637$\,s
& 0.18 & \secondcell{4.37} & \secondcell{5.11}
& \secondcell{0.07} & 25.41 & 27.33
& \bestcell{0.04} & \bestcell{0.31} & \bestcell{1.07} \\
& $+0.283$\,s
& \secondcell{0.22} & \secondcell{4.62} & \secondcell{5.53}
& 1.21 & 19.00 & 23.24
& \bestcell{0.09} & \bestcell{0.43} & \bestcell{1.15} \\
& $+0.912$\,s
& \secondcell{0.12} & 4.13 & \secondcell{5.21}
& 0.19 & \secondcell{1.12} & 8.68
& \bestcell{0.02} & \bestcell{0.81} & \bestcell{1.29} \\
\midrule
\multirow{3}{*}{Baseline}
& 5\,cm
& \secondcell{0.26} & \secondcell{4.16} & \secondcell{5.06}
& 1.93 & 8.97 & 10.63
& \bestcell{0.08} & \bestcell{0.49} & \bestcell{0.77} \\
& 20\,cm
& \secondcell{0.18} & \secondcell{3.30} & \secondcell{3.94}
& 2.41 & 11.22 & 12.55
& \bestcell{0.13} & \bestcell{0.58} & \bestcell{0.60} \\
& 50\,cm
& \secondcell{0.37} & \secondcell{3.71} & \secondcell{4.50}
& 0.93 & 5.63 & 6.49
& \bestcell{0.06} & \bestcell{0.66} & \bestcell{0.70} \\
\midrule
\multirow{3}{*}{Rotation}
& $5^\circ$
& \secondcell{0.40} & \secondcell{3.98} & \secondcell{4.85}
& 2.60 & 25.49 & 35.79
& \bestcell{0.10} & \bestcell{1.26} & \bestcell{1.30} \\
& $10^\circ$
& \secondcell{0.09} & \secondcell{3.63} & \secondcell{4.09}
& 7.10 & 31.85 & 35.12
& \bestcell{0.02} & \bestcell{1.47} & \bestcell{1.46} \\
& $20^\circ$
& \secondcell{0.19} & \bestcell{1.57} & \bestcell{1.62}
& 3.65 & 12.35 & 18.59
& \bestcell{0.12} & \secondcell{1.61} & \secondcell{1.63} \\
\midrule
\multicolumn{2}{l}{Median over the 9 common cases}
& \secondcell{0.19} & \secondcell{3.98} & \secondcell{4.85}
& 1.93 & 12.35 & 18.59
& \bestcell{0.08} & \bestcell{0.66} & \bestcell{1.15} \\
\multicolumn{2}{l}{Mean over the 9 common cases}
& \secondcell{0.22} & \secondcell{3.72} & \secondcell{4.43}
& 2.23 & 15.67 & 19.82
& \bestcell{0.07} & \bestcell{0.85} & \bestcell{1.11} \\
\bottomrule
\end{tabular}
\vspace{-5pt}
\end{table*}

\begin{table*}[t]
\centering
\caption{\textbf{Event-camera calibration on real data}: mean $\pm$ sample standard deviation over the trials. We report two approximate baselines using the same event camera paired with a Nikon RGB camera; the recordings are not synchronized. Each method is run on 40 random draws of the training frames per setting. Est.\ baseline is the estimated distance between the two cameras. The distortion coefficients are listed in \cref{tab:calib_real_full}.}
\vspace{-5pt}
\label{tab:calib_real}
\footnotesize
\setlength{\tabcolsep}{5pt}
\begin{tabular}{ll cccc c}
\toprule
Baseline & Method
& $f_x$ (px) & $f_y$ (px)
& $c_x$ (px) & $c_y$ (px)
& Est.\ baseline (cm) \\
\midrule
\multirow{2}{*}{\shortstack[l]{$\approx10$\,cm}}
& E2Calib~\cite{muglikar2021calibrate}
& $1228.60 \pm 8.41$ & $1228.28 \pm 8.71$
& $338.38 \pm 9.76$ & $249.74 \pm 6.63$
& $15.47 \pm 1.36$ \\
& \method (ours)
& $1231.47 \pm 15.38$ & $1229.94 \pm 14.66$
& $322.07 \pm 1.09$ & $252.53 \pm 2.70$
& $12.08 \pm 0.51$ \\
\midrule
\multirow{2}{*}{\shortstack[l]{$\approx25$\,cm}}
& E2Calib~\cite{muglikar2021calibrate}
& $1255.49 \pm 11.21$ & $1252.46 \pm 10.88$
& $337.00 \pm 8.36$ & $246.40 \pm 10.78$
& $32.31 \pm 2.13$ \\
& \method (ours)
& $1196.91 \pm 51.07$ & $1190.28 \pm 62.76$
& $319.63 \pm 5.26$ & $254.99 \pm 4.66$
& $26.39 \pm 1.17$ \\
\bottomrule
\end{tabular}
\end{table*}

\noindent \textbf{Synthetic data.}
We render recordings in Blender and simulate noise-free events with an ESIM-style simulator~\cite{Rebecq18corl}.
A 60\,fps RGB camera ($1920 \times 1080$) and an event camera ($640 \times 512$), both without distortion, observe a checkerboard moved at a distance of about 3\,m.
From a default configuration (20\,cm baseline, parallel optical axes, no temporal offset), we vary one factor at a time: the temporal offset $\dt \in \{-0.637, +0.283, +0.912\}$\,s, the baseline (5, 20, and 50\,cm), and the rotation of the event camera toward the RGB camera ($5^\circ$, $10^\circ$, and $20^\circ$).

\noindent \textbf{Real data.}
We mount an event camera (LUCID Triton2 EVS, $640\times512$) and a Nikon COOLPIX P1100 ($1920\times1080$, 59.94\,fps, rolling shutter, auto-exposure) on a rigid rig without a shared trigger, at baselines of approximately 10 and 25\,cm, and record a hand-held checkerboard; the recording offset amounts to several seconds.
An Intel RealSense D457 depth camera registered to the RGB camera provides the ground truth for depth.

\noindent \textbf{Baselines.}
For the event-camera intrinsics, we compare with E2Calib~\cite{muglikar2021calibrate}, which calibrates on images reconstructed from the events with E2VID~\cite{rebecq2019high}, and Corner2Tag~\cite{ryu2026corner2tag}, which detects the checkerboard corners in images of accumulated events.
As its code was not public at the time of writing, we use our reimplementation of Corner2Tag, whose results may differ from the original (\cref{sec:supp_corner2tag}).
For the temporal offset and the extrinsics, we compare with EF-Calib~\cite{wang2024ef}, which estimates them jointly with the intrinsics from a dedicated pattern, and with E2Calib, which assumes synchronized recordings and therefore receives the offset $\dtpre$ of our temporal pre-alignment (\cref{sec:prealign}).

\noindent \textbf{Metrics.}
On synthetic data, we report errors with respect to the ground truth: the relative focal-length error $e_f$, the principal-point error $e_c$, and the projection error $e_{\pi}$, i.e., the mean pixel distance between the projections of a grid of viewing rays under the estimated and the ground-truth camera model, which covers intrinsics and distortion jointly; the rotation error $e_{R}$, the translation error $e_{t}$, and the baseline-length error of the extrinsics; and the absolute error $e_{\dt}$ of the temporal offset.
On real data, without ground truth, we repeat the calibration on 40 random subsets of 24 frames of each recording, the same for all methods, and report the mean and the standard deviation of the estimates; as the event camera is the same in all settings, its intrinsics should agree across trials and settings.
For depth, we report the RMSE, the MAE, and the absolute relative error (AbsRel) on the moving regions, without rescaling any prediction to the ground truth.

\noindent \textbf{Implementation details.}
All learned components are used with their public weights and kept frozen, and the parameter values are the same for all recordings (\cref{sec:supp_impl,sec:supp_exp} list all values and the full protocols).

\subsection{Event-Camera Intrinsics}
\label{sec:exp_intrinsics}

\begin{figure*}[t]
    \centering
    \includegraphics[width=0.85\linewidth]{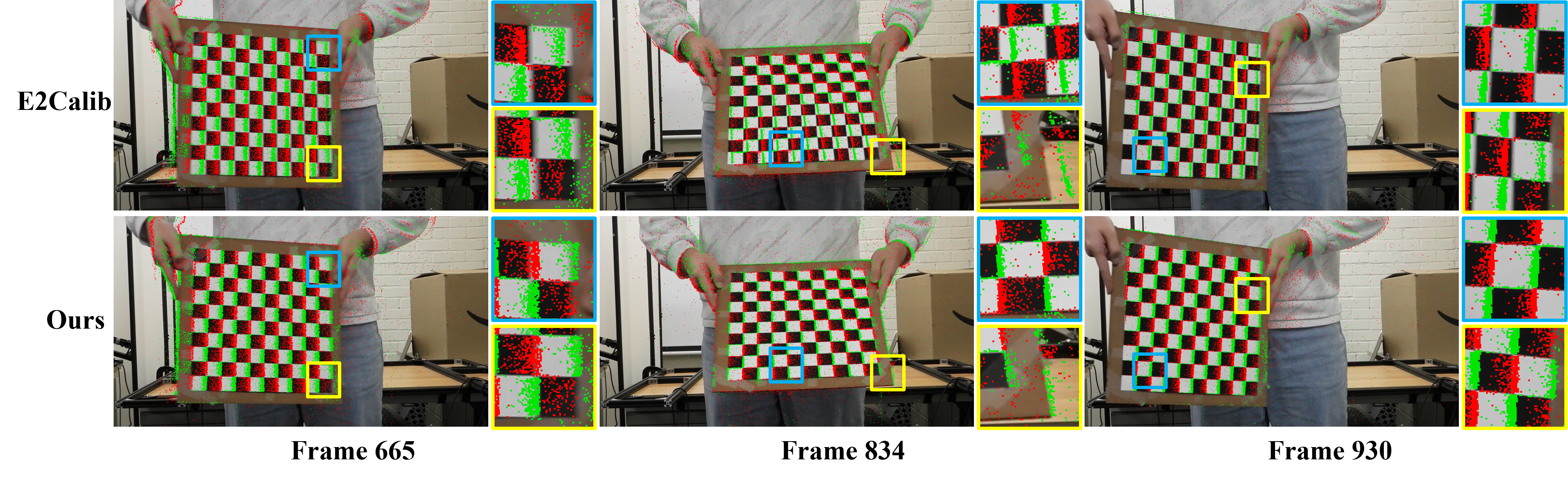}
    \vspace{-10pt}
    \caption{\textbf{Qualitative comparison on real data}. Events are warped into RGB frames with the intrinsics, extrinsics, and temporal offset estimated by each method; red and green denote the two polarities, and the insets enlarge the marked regions. With \method, the events coincide with the boundaries between the squares.}
    \label{fig:warpcompare}
    \vspace{-10pt}
\end{figure*}

\noindent \textbf{Synthetic data.}
In all nine configurations of \cref{tab:calib_syn}, \method estimates the focal length to within $0.13\%$ and the principal point to within $1.61$\,px, and its projection error stays below $1.63$\,px.
Its median errors are $0.08\%$, $0.66$\,px, and $1.15$\,px, compared with $0.19\%$, $3.98$\,px, and $4.85$\,px for E2Calib and $1.93\%$, $12.35$\,px, and $18.59$\,px for our reimplementation of Corner2Tag.
\method has the lowest error in every configuration except for the rotation of $20^\circ$, where E2Calib is on par in the principal point and the projection error.
Its focal length and principal point are as accurate with a recording offset or a different baseline as in the default configuration; the principal-point error grows mildly with the relative rotation, from $1.26$\,px at $5^\circ$ to $1.61$\,px at $20^\circ$, and a recording offset raises the projection error from $0.60$\,px to $1.07$--$1.29$\,px through a small error in the radial distortion (\cref{tab:calib_syn_full}).

\noindent \textbf{Real data.}
\Cref{tab:calib_real} reports the statistics over the repeated trials on the real recordings.
\method yields a more repeatable principal point than E2Calib, with a standard deviation of $1.09$--$5.26$\,px in $c_x$ against $8.36$--$9.76$\,px.
Its focal length, in contrast, varies more between trials ($1.2$--$5.3\%$ against $0.7$--$0.9\%$), and so do its radial distortion coefficients (\cref{tab:calib_real_full}).
With only 24 frames per trial, the depth of the board varies by less than $0.25$\,m, so the focal length is the least constrained parameter and a trial can settle in a local minimum of the periodic event-to-line objective; the full recording, as used in \cref{tab:calib_syn}, covers a wider range of depths and orientations (\cref{sec:supp_exp}).

\subsection{Temporal Offset and Extrinsics}
\label{sec:exp_cocalib}

\noindent \textbf{Synthetic data.}
\Cref{tab:cocalib_syn} evaluates the temporal offset and the extrinsics of the pair on temporally misaligned synthetic recordings.
As EF-Calib refines the temporal offset starting from zero, we use three offsets below $50$\,ms, $\Delta t\in\{-0.037,+0.013,+0.044\}$\,s.
\method recovers the three offsets with an error of $0.38$--$0.41$\,ms, i.e., less than $2.5\%$ of the frame period, a rotation error below $0.1^\circ$, and a translation error below $2$\,mm for a baseline of $200$\,mm.
E2Calib, which receives the offset of our temporal pre-alignment (errors of $2.3$--$7.9$\,ms), has rotation errors of $0.35$--$0.41^\circ$ and translation errors of $3.2$--$7.5$\,mm, while its baseline length is as accurate as ours (errors below $0.7$\,mm for both methods).
EF-Calib estimates the offset to within $4$\,ms, but its translation is off by $14$--$21$\,mm, mostly along the baseline direction (baseline-length errors of $12$--$20$\,mm).

\begin{table}[t]
\centering
\caption{Event--RGB co-calibration on synthetic data with temporal offsets $\Delta t$. We report the estimated temporal offset and the rotation, translation, and baseline errors of the extrinsics. EF-Calib is evaluated on sequences rendered with its own pattern along the same trajectory and with the same offsets. Within each case, dark and light blue mark the best and the second-best result for each metric; estimated offsets are ranked by their absolute errors with respect to the ground truth.}
\label{tab:cocalib_syn}
\small
\setlength{\tabcolsep}{3pt}
\resizebox{\linewidth}{!}{%
\begin{tabular}{ll cccc}
\toprule
Offset & Method & Est. Offset (s)
& Rot. ($^\circ$) $\downarrow$
& Trans. (mm) $\downarrow$
& Baseline (mm) $\downarrow$ \\
\midrule
\multirow{3}{*}{$-0.037$\,s}
& E2Calib~\cite{muglikar2021calibrate}
& \secondcell{-0.034749}
& \secondcell{0.367}
& \secondcell{3.674}
& \bestcell{0.467} \\
& EF-Calib~\cite{wang2024ef}
& -0.040765
& 0.370
& 18.080
& 15.660 \\
& \method (ours)
& \bestcell{-0.037381}
& \bestcell{0.084}
& \bestcell{1.179}
& \secondcell{0.623} \\
\midrule
\multirow{3}{*}{$+0.013$\,s}
& E2Calib~\cite{muglikar2021calibrate}
& 0.018580
& 0.406
& \secondcell{7.515}
& \secondcell{0.501} \\
& EF-Calib~\cite{wang2024ef}
& \secondcell{0.013503}
& \secondcell{0.260}
& 13.640
& 11.980 \\
& \method (ours)
& \bestcell{0.012614}
& \bestcell{0.051}
& \bestcell{1.879}
& \bestcell{0.441} \\
\midrule
\multirow{3}{*}{$+0.044$\,s}
& E2Calib~\cite{muglikar2021calibrate}
& 0.051916
& \secondcell{0.346}
& \secondcell{3.172}
& \bestcell{0.313} \\
& EF-Calib~\cite{wang2024ef}
& \secondcell{0.041771}
& 0.480
& 21.420
& 19.860 \\
& \method (ours)
& \bestcell{0.043588}
& \bestcell{0.065}
& \bestcell{1.329}
& \secondcell{0.580} \\
\bottomrule
\end{tabular}}
\vspace{-10pt}
\end{table}

\noindent \textbf{Real data.}
On the real recordings, neither the temporal offset nor the extrinsics have a reference, so we check them in two indirect ways.
First, we compare the estimated baseline (\cref{tab:calib_real}) with the distance between the centre lines of the two cameras measured with a tape, about $10$ and $25$\,cm.
\method estimates $12.08$ and $26.39$\,cm, with a standard deviation of $0.51$--$1.17$\,cm over the trials, i.e., within $1.4$--$2.1$\,cm of the tape measurement, whereas E2Calib gives $15.47$ and $32.31$\,cm (standard deviation of $1.36$--$2.13$\,cm), i.e., $5.5$--$7.3$\,cm off.
Second, \cref{fig:warpcompare} warps the events on the board into the RGB frames with the calibration of each method, through the board plane given by the RGB pose at the time of each event; E2Calib uses the offset of our temporal pre-alignment.
With \method, the events fall on the boundaries between the squares, where a moving checkerboard triggers them, whereas with E2Calib they are visibly displaced.

\subsection{Dynamic 3D Geometry and Applications}
\label{sec:exp_fourd}

\begin{figure}[t]
    \centering
    \includegraphics[width=\linewidth]{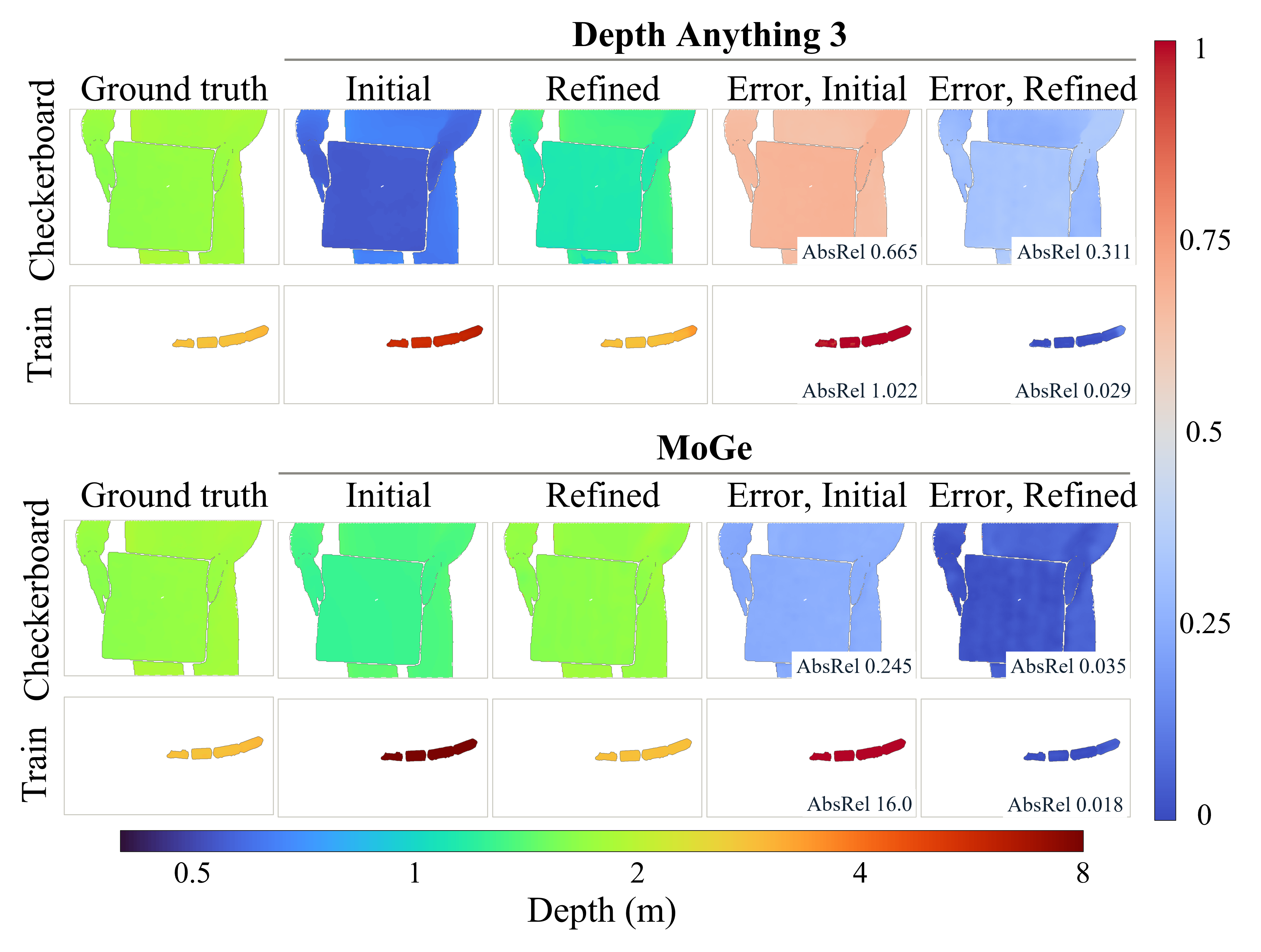}
    \vspace{-10pt}
    \caption{\textbf{Event-guided depth refinement on real and synthetic data.} The first row shows a frame of the recording of \cref{tab:depth_prior}, the second row shows Blender synthetic data, each with DA3 and MoGe. From left to right: RGB frame, ground-truth depth, initial depth of the model, refined depth, and the absolute relative errors of the initial and the refined depth with respect to the ground-truth depth. Only the moving region is shown.}
    \label{fig:depthoptimization}
    \vspace{-10pt}
\end{figure}

\noindent \textbf{Event-guided metric depth.}
We evaluate the event-guided depth of \cref{sec:fourd} on a real recording with a moving person and board against the depth camera, with three 3D foundation models as the prior: Depth Anything~3 (DA3)~\cite{lin2511depth}, MoGe~\cite{kong2026moge}, and VGGT-$\Omega$~\cite{wang2026vggt} (\cref{tab:depth_prior,fig:depthoptimization}).
The refinement reduces the error for both models that predict metric depth: the AbsRel drops from $0.222$ to $0.038$ for MoGe and from $0.636$ to $0.442$ for DA3.
As \cref{fig:depthoptimization} shows, the error of the initial depth is nearly constant over the moving region, i.e., the models recover the shape of the scene but not its metric scale, which the events observed from the second viewpoint constrain.
For VGGT-$\Omega$, which predicts depth only up to scale, the refinement recovers metric depth with an AbsRel of $0.090$.

\noindent \textbf{Dynamic 3D reconstruction and slow motion.}
Building on this depth, the temporal geometry refinement and the sub-frame trajectories of \cref{sec:fourd} yield dynamic 3D reconstructions and 4D slow motion.
These results, on real and synthetic scenes, can be found in \cref{sec:supp_more} and in the accompanying videos.

\begin{table}[t]
\centering
\footnotesize
\setlength{\tabcolsep}{2.6pt}
\caption{\textbf{Event-guided depth refinement with different 3D foundation models} on a real recording with depth-camera ground truth: errors of the predicted (\emph{Initial}) and the refined depth (Sec.~\ref{sec:fourd}) on the moving region, without rescaling to the ground truth; RMSE and MAE in metres. VGGT-$\Omega$ predicts depth only up to scale, so its initial depth has no metric error.}
\label{tab:depth_prior}
\begin{tabular}{@{}l ccc ccc@{}}
\toprule
 & \multicolumn{3}{c}{Initial} & \multicolumn{3}{c}{Refined} \\
\cmidrule(lr){2-4} \cmidrule(l){5-7}
Model & RMSE$\downarrow$ & MAE$\downarrow$ & AbsRel$\downarrow$ & RMSE$\downarrow$ & MAE$\downarrow$ & AbsRel$\downarrow$\\
\midrule
DA3 & 1.071 & 1.068 & 0.636 & \textbf{0.764} & \textbf{0.741} & \textbf{0.442}\\
MoGe & 0.381 & 0.374 & 0.222 & \textbf{0.117} & \textbf{0.065} & \textbf{0.038}\\
VGGT-$\Omega$ & -- & -- & -- & 0.256 & 0.151 & 0.090\\
\bottomrule
\end{tabular}
\vspace{-10pt}
\end{table}

\subsection{Ablation Studies}
\label{sec:exp_ablation}

\noindent \textbf{Calibration components.}
\Cref{tab:ablation_calib} isolates the contribution of each stage of \cref{sec:prealign,sec:calib} on the synthetic recordings with temporal offsets.
In the upper part, the paired-track score yields an integer lag with an error of $8.45$\,ms, i.e., half of the frame period; the epipolar--temporal refinement reduces it to $3.86$\,ms, and the residual offset estimated from the raw events to $0.40$\,ms.
In the lower part, this last step and the board motion matter most.
Without the residual offset, the remaining $3.86$\,ms, a quarter of a frame, increase the projection error from $1.18$ to $6.09$\,px and the translation error from $2.24$ to $51.34$\,mm; without the board motion within the event window, the residual offset is not observable and the errors grow to $9.51$\,px and $120.31$\,mm, which is no better than the coarse initialization ($10.19$\,px and $46.48$\,mm).
Removing the correspondence term or the staged optimization changes the offset error by less than $0.01$\,ms and the projection error by at most $0.3$\,px, while the focal-length and translation errors increase from $0.05\%$ to $0.08$--$0.13\%$ and from $2.24$\,mm to $3.05$--$4.98$\,mm; the full method has the lowest error in three of the five metrics and is close to the best variant in the other two.

\begin{table}[t]
\centering
\caption{\textbf{Ablation of the calibration} on synthetic recordings with temporal offsets $-0.637$, $+0.283$, and $+0.912$\,s. We report the mean error over the three recordings. The upper part follows the temporal offset through the stages of the method; the lower part evaluates coarse initialization and removes individual components from the full method. Dark and light blue mark the best and second-best results for each metric, respectively, with ties highlighted equally.}
\label{tab:ablation_calib}
\small
\setlength{\tabcolsep}{3pt}
\resizebox{\linewidth}{!}{%
\begin{tabular}{l ccccc}
\toprule
Variant
& $e_{\dt}$ (ms) $\downarrow$
& $e_f$ (\%) $\downarrow$
& $e_{\pi}$ (px) $\downarrow$
& $e_R$ ($^\circ$) $\downarrow$
& $e_t$ (mm) $\downarrow$ \\
\midrule
Paired-track score, $\lag^{(1)}$
& 8.445 & n/a & n/a & n/a & n/a \\
+ epipolar--temporal refinement, $\dtpre$
& 3.860 & n/a & n/a & n/a & n/a \\
Full \method{}
& 0.404
& \bestcell{0.049}
& 1.177
& \bestcell{0.062}
& \bestcell{2.242} \\
\midrule
Coarse initialization only, $\Par^{(0)}$
& 5.470 & 1.336 & 10.191 & 0.880 & 46.476 \\
w/o correspondence term ($\beta = 0$)
& \bestcell{0.396}
& 0.134
& \bestcell{0.876}
& \secondcell{0.065}
& 4.984 \\
w/o board motion ($\dot{\Bd}_k = \mathbf{0}$)
& 5.470 & 3.693 & 9.508 & 0.216 & 120.308 \\
w/o residual offset ($\res$ fixed)
& 3.860 & 1.559 & 6.089 & 0.170 & 51.340 \\
w/o staged optimization
& \secondcell{0.402}
& \secondcell{0.078}
& \secondcell{1.107}
& 0.071
& \secondcell{3.045} \\
Full \method{}
& 0.404
& \bestcell{0.049}
& 1.177
& \bestcell{0.062}
& \bestcell{2.242} \\
\bottomrule
\end{tabular}}
\end{table}

\section{Conclusion}
\label{sec:conclusion}
We presented \method, a spatiotemporal calibration method for event--RGB cameras that record without a shared trigger or clock, which detects the calibration target only in the RGB camera and requires no task-specific training.
Its temporal pre-alignment recovers the recording offset from the motion observed in both views and from the epipolar geometry between them, and its raw-event calibration fits individual events to the grid lines of a checkerboard that moves in continuous time, which jointly estimates the event-camera parameters, the metric extrinsics, and a residual temporal offset.
Building on the calibrated rig, we use sparse event constraints to correct the scale and local errors of a monocular depth prior and refine moving-object geometry along pretrained tracks.

\vspace{3pt}
\noindent \textbf{Limitations.}
The method relies on board poses estimated by PnP in the RGB frames, so heavily blurred frames do not contribute to the calibration.
It models the timing discrepancy as a constant offset: clock drift, exposure changes, and the rolling shutter of the RGB camera are bounded but not estimated.
The geometry refinement currently assumes a static rig and retains the learned depth prior where event constraints are weak.
Extending the calibration to natural scenes without a target and the reconstruction to moving rigs are promising directions for future work.

\bibliographystyle{ACM-Reference-Format}
\bibliography{main}

@String(CVPR= {IEEE Conf. Comput. Vis. Pattern Recog.})

@String(ICCV= {Int. Conf. Comput. Vis.})

@String(CVPRW= {IEEE Conf. Comput. Vis. Pattern Recog. Worksh.})

@String(CVPR  = {CVPR})

@String(ICCV  = {ICCV})

@String(CVPRW= {CVPRW})

@inproceedings{pan2019bringing,
  title={Bringing a blurry frame alive at high frame-rate with an event camera},
  author={Pan, Liyuan and Scheerlinck, Cedric and Yu, Xin and Hartley, Richard and Liu, Miaomiao and Dai, Yuchao},
  booktitle={2019 IEEE/CVF Conference on Computer Vision and Pattern Recognition (CVPR)},
  pages={6813--6822},
  year={2019},
  organization={IEEE}
}

@article{gehrig2021dsec,
  title={Dsec: A stereo event camera dataset for driving scenarios},
  author={Gehrig, Mathias and Aarents, Willem and Gehrig, Daniel and Scaramuzza, Davide},
  journal={IEEE Robotics and Automation Letters},
  volume={6},
  number={3},
  pages={4947--4954},
  year={2021},
  publisher={IEEE}
}

@article{zhang2000flexible,
  title={A flexible new technique for camera calibration},
  author={Zhang, Zhengyou},
  journal={IEEE Transactions on pattern analysis and machine intelligence},
  volume={22},
  number={11},
  pages={1330--1334},
  year={2000},
  publisher={Ieee}
}

@inproceedings{muglikar2021calibrate,
  title={How to calibrate your event camera},
  author={Muglikar, Manasi and Gehrig, Mathias and Gehrig, Daniel and Scaramuzza, Davide},
  booktitle={2021 IEEE/CVF Conference on Computer Vision and Pattern Recognition Workshops (CVPRW)},
  pages={1403--1409},
  year={2021},
  organization={IEEE}
}

@article{rebecq2019high,
  title={High speed and high dynamic range video with an event camera},
  author={Rebecq, Henri and Ranftl, Ren{\'e} and Koltun, Vladlen and Scaramuzza, Davide},
  journal={IEEE transactions on pattern analysis and machine intelligence},
  volume={43},
  number={6},
  pages={1964--1980},
  year={2019},
  publisher={IEEE}
}

@article{wang2024motion,
  title={Motion-error-free calibration of event camera systems using a flashing target},
  author={Wang, Yongqing and He, Shiyu and Fei, Yufan and Liu, Xingjian},
  journal={Optics express},
  volume={32},
  number={15},
  pages={26833--26845},
  year={2024},
  publisher={Optica Publishing Group}
}

@inproceedings{huang2021dynamic,
  title={Dynamic event camera calibration},
  author={Huang, Kun and Wang, Yifu and Kneip, Laurent},
  booktitle={2021 IEEE/RSJ International Conference on Intelligent Robots and Systems (IROS)},
  pages={7021--7028},
  year={2021},
  organization={IEEE}
}

@article{salah2024calib,
  title={E-Calib: A fast, robust, and accurate calibration toolbox for event cameras},
  author={Salah, Mohammed and Ayyad, Abdulla and Humais, Muhammad and Gehrig, Daniel and Abusafieh, Abdelqader and Seneviratne, Lakmal and Scaramuzza, Davide and Zweiri, Yahya},
  journal={IEEE Transactions on Image Processing},
  volume={33},
  pages={3977--3990},
  year={2024},
  publisher={IEEE}
}

@article{chen2025ekalibr,
  title={eKalibr: Dynamic intrinsic calibration for event cameras from first principles of events},
  author={Chen, Shuolong and Li, Xingxing and Yuan, Liu and Liu, Ziao},
  journal={IEEE Robotics and Automation Letters},
  volume={10},
  number={7},
  pages={7094--7101},
  year={2025},
  publisher={IEEE}
}

@InProceedings{ryu2026corner2tag,
  author    = {Taehun Ryu and Changwoo Kang and Kyungdon Joo},
  title     = {From Corners to Fiducial Tags: Revisiting Checkerboard Calibration for Event Cameras},
  booktitle = {IEEE/CVF Conference on Computer Vision and Pattern Recognition (CVPR)},
  year      = {2026}
}

@inproceedings{furgale2013unified,
  title={Unified temporal and spatial calibration for multi-sensor systems},
  author={Furgale, Paul and Rehder, Joern and Siegwart, Roland},
  booktitle={2013 IEEE/RSJ International Conference on Intelligent Robots and Systems},
  pages={1280--1286},
  year={2013},
  organization={IEEE}
}

@article{rehder2016general,
  title={A general approach to spatiotemporal calibration in multisensor systems},
  author={Rehder, Joern and Siegwart, Roland and Furgale, Paul},
  journal={IEEE Transactions on Robotics},
  volume={32},
  number={2},
  pages={383--398},
  year={2016},
  publisher={IEEE}
}

@article{wang2024ef,
  title={EF-Calib: Spatiotemporal calibration of event-and frame-based cameras using continuous-time trajectories},
  author={Wang, Shaoan and Xin, Zhanhua and Hu, Yaoqing and Li, Dongyue and Zhu, Mingzhu and Yu, Junzhi},
  journal={IEEE Robotics and Automation Letters},
  volume={9},
  number={11},
  pages={10280--10287},
  year={2024},
  publisher={IEEE}
}

@article{chen2025ekalibrstereo,
  title={eKalibr-Stereo: Continuous-time spatiotemporal calibration for event-based stereo visual systems},
  author={Chen, Shuolong and Li, Xingxing and Yuan, Liu},
  journal={IEEE Robotics and Automation Letters},
  year={2025},
  publisher={IEEE}
}

@inproceedings{gossard2024ewand,
  title={ewand: An extrinsic calibration framework for wide baseline frame-based and event-based camera systems},
  author={Gossard, Thomas and Ziegler, Andreas and Kolmar, Levin and Tebbe, Jonas and Zell, Andreas},
  booktitle={2024 IEEE International Conference on Robotics and Automation (ICRA)},
  pages={14534--14540},
  year={2024},
  organization={IEEE}
}

@article{gallego2020event,
  title={Event-based vision: A survey},
  author={Gallego, Guillermo and Delbr{\"u}ck, Tobi and Orchard, Garrick and Bartolozzi, Chiara and Taba, Brian and Censi, Andrea and Leutenegger, Stefan and Davison, Andrew J and Conradt, J{\"o}rg and Daniilidis, Kostas and others},
  journal={IEEE transactions on pattern analysis and machine intelligence},
  volume={44},
  number={1},
  pages={154--180},
  year={2020},
  publisher={IEEE}
}

@inproceedings{sundar2024generalized,
  title={Generalized event cameras},
  author={Sundar, Varun and Dutson, Matthew and Ardelean, Andrei and Bruschini, Claudio and Charbon, Edoardo and Gupta, Mohit},
  booktitle={2024 IEEE/CVF Conference on Computer Vision and Pattern Recognition (CVPR)},
  pages={25007--25017},
  year={2024},
  organization={IEEE}
}

@article{cai2024accurate,
  title={Accurate event camera calibration with fourier transform},
  author={Cai, Bolin and Zi, Ami and Yang, Jun and Li, Guoliang and Zhang, Yang and Wu, Qiujie and Tong, Chenen and Liu, Wenxiang and Chen, Xiangcheng},
  journal={IEEE Transactions on Instrumentation and Measurement},
  volume={73},
  pages={1--12},
  year={2024},
  publisher={IEEE}
}

@article{zhang2022formulating,
  title={Formulating event-based image reconstruction as a linear inverse problem with deep regularization using optical flow},
  author={Zhang, Zelin and Yezzi, Anthony J and Gallego, Guillermo},
  journal={IEEE transactions on pattern analysis and machine intelligence},
  volume={45},
  number={7},
  pages={8372--8389},
  year={2022},
  publisher={IEEE}
}

@article{song2025camera,
  title={Camera calibration via circular patterns: A comprehensive framework with measurement uncertainty and unbiased projection model},
  author={Song, Chaehyeon and Lee, Dongjae and Lim, Jongwoo and Kim, Ayoung},
  journal={arXiv e-prints},
  pages={arXiv--2506},
  year={2025}
}

@inproceedings{gallego2018unifying,
  title={A unifying contrast maximization framework for event cameras, with applications to motion, depth, and optical flow estimation},
  author={Gallego, Guillermo and Rebecq, Henri and Scaramuzza, Davide},
  booktitle={2018 IEEE/CVF Conference on Computer Vision and Pattern Recognition},
  pages={3867--3876},
  year={2018},
  organization={IEEE}
}

@inproceedings{sommer2020efficient,
  title={Efficient derivative computation for cumulative b-splines on lie groups},
  author={Sommer, Christiane and Usenko, Vladyslav and Schubert, David and Demmel, Nikolaus and Cremers, Daniel},
  booktitle={2020 IEEE/CVF Conference on Computer Vision and Pattern Recognition (CVPR)},
  pages={11145--11153},
  year={2020},
  organization={IEEE}
}

@article{mueggler2018continuous,
  title={Continuous-time visual-inertial odometry for event cameras},
  author={Mueggler, Elias and Gallego, Guillermo and Rebecq, Henri and Scaramuzza, Davide},
  journal={IEEE Transactions on Robotics},
  volume={34},
  number={6},
  pages={1425--1440},
  year={2018},
  publisher={IEEE}
}

@article{caspi2002spatio,
  title={Spatio-temporal alignment of sequences},
  author={Caspi, Yaron and Irani, Michal},
  journal={IEEE Transactions on Pattern Analysis and Machine Intelligence},
  volume={24},
  number={11},
  pages={1409--1424},
  year={2002},
  publisher={IEEE}
}

@article{padua2008linear,
  title={Linear sequence-to-sequence alignment},
  author={Padua, Flavio and Carceroni, Rodrigo and Santos, Geraldo and Kutulakos, Kiriakos},
  journal={IEEE Transactions on Pattern Analysis and Machine Intelligence},
  volume={32},
  number={2},
  pages={304--320},
  year={2008},
  publisher={IEEE}
}

@inproceedings{karaev2025cotracker3,
  title={CoTracker3: Simpler and Better Point Tracking by Pseudo-Labeling Real Videos},
  author={Karaev, Nikita and Makarov, Yuri and Wang, Jianyuan and Neverova, Natalia and Vedaldi, Andrea and Rupprecht, Christian},
  booktitle={2025 IEEE/CVF International Conference on Computer Vision (ICCV)},
  pages={1--10},
  year={2025},
  organization={IEEE}
}

@inproceedings{detone2018superpoint,
  title={Superpoint: Self-supervised interest point detection and description},
  author={DeTone, Daniel and Malisiewicz, Tomasz and Rabinovich, Andrew},
  booktitle={2018 IEEE/CVF conference on computer vision and pattern recognition workshops (CVPRW)},
  pages={337--33712},
  year={2018},
  organization={IEEE}
}

@inproceedings{lindenberger2023lightglue,
  title={Lightglue: Local feature matching at light speed},
  author={Lindenberger, Philipp and Sarlin, Paul-Edouard and Pollefeys, Marc},
  booktitle={2023 IEEE/CVF international conference on computer vision (ICCV)},
  pages={17581--17592},
  year={2023},
  organization={IEEE}
}

@book{hartley2003multiple,
  title={Multiple view geometry in computer vision},
  author={Hartley, Richard and Zisserman, Andrew and others},
  volume={2},
  number={3},
  year={2003},
  publisher={Cambridge university press, Cambridge}
}

@article{lin2511depth,
  title={Depth Anything 3: Recovering the visual space from any views,(2025)},
  author={Lin, Haotong and Chen, Sili and Liew, Junhao and Chen, Donny Y and Li, Zhenyu and Shi, Guang and Feng, Jiashi and Kang, Bingyi},
  journal={arXiv preprint arXiv:2511.10647},
  volume={5}
}

@inproceedings{jin2025stereo4d,
  title={Stereo4d: Learning how things move in 3d from internet stereo videos},
  author={Jin, Linyi and Tucker, Richard and Li, Zhengqi and Fouhey, David and Snavely, Noah and Holynski, Aleksander},
  booktitle={Proceedings of the IEEE/CVF Conference on Computer Vision and Pattern Recognition},
  pages={10497--10509},
  year={2025}
}

@inproceedings{besl1992method,
  title={Method for registration of 3-D shapes},
  author={Besl, Paul J and McKay, Neil D},
  booktitle={Sensor fusion IV: control paradigms and data structures},
  volume={1611},
  pages={586--606},
  year={1992},
  organization={Spie}
}

@inproceedings{teed2020raft,
  title={Raft: Recurrent all-pairs field transforms for optical flow},
  author={Teed, Zachary and Deng, Jia},
  booktitle={European conference on computer vision},
  pages={402--419},
  year={2020},
  organization={Springer}
}

@inproceedings{carion2026sam,
  title={Sam 3: Segment anything with concepts},
  author={Carion, Nicolas and Gustafson, Laura and Hu, Yuan-Ting and Debnath, Shoubhik and Hu, Ronghang and Suris Coll-Vinent, Didac and Ryali, Chaitanya and Alwala, Kalyan Vasudev and Khedr, Haitham and Huang, Andrew and others},
  booktitle={International conference on learning representations},
  volume={2026},
  pages={138846--138923},
  year={2026}
}

@inproceedings{tulyakov2021time,
  title={Time lens: Event-based video frame interpolation},
  author={Tulyakov, Stepan and Gehrig, Daniel and Georgoulis, Stamatios and Erbach, Julius and Gehrig, Mathias and Li, Yuanyou and Scaramuzza, Davide},
  booktitle={2021 IEEE/CVF Conference on Computer Vision and Pattern Recognition (CVPR)},
  pages={16150--16159},
  year={2021},
  organization={IEEE}
}

@inproceedings{ma2024timelens,
  title={Timelens-xl: Real-time event-based video frame interpolation with large motion},
  author={Ma, Yongrui and Guo, Shi and Chen, Yutian and Xue, Tianfan and Gu, Jinwei},
  booktitle={European Conference on Computer Vision},
  pages={178--194},
  year={2024},
  organization={Springer}
}

@article{kong2026moge,
  title={MoGe-3: Fine-detail monocular geometry estimation with self-guided sparse volumetric refinement},
  author={Kong, Lingyu and Li, Ruicheng and Wang, Ruicheng and Xu, Sicheng and Yao, Chengtang and Xiang, Jianfeng and Yang, Jiaolong},
  journal={arXiv e-prints},
  pages={arXiv--2607},
  year={2026}
}

@article{wang2026vggt,
  title={VGGT-$Omega$},
  author={Wang, Jianyuan and Chen, Minghao and Zhang, Shangzhan and Karaev, Nikita and Sch{\"o}nberger, Johannes and Labatut, Patrick and Bojanowski, Piotr and Novotny, David and Vedaldi, Andrea and Rupprecht, Christian},
  journal={arXiv preprint arXiv:2605.15195},
  year={2026}
}

@Article{Rebecq18corl,
  author        = {Henri Rebecq and Daniel Gehrig and Davide Scaramuzza},
  title         = {{ESIM}: an Open Event Camera Simulator},
  journal       = {Conf. on Robotics Learning (CoRL)},
  year          = 2018,
  month         = oct
}

@article{hessenthaler2026simplified,
  title={Simplified Cross-Modal Calibration for Heterogeneous Event-RGB Stereo Systems},
  author={Hessenthaler, Nico and M{\"u}ller, Adam T and Stache, Nicolaj C},
  journal={arXiv preprint arXiv:2608.22965},
  year={2026}
}

@article{wu2025eag3r,
  title={EAG3R: Event-augmented 3D geometry estimation for dynamic and extreme-lighting scenes},
  author={Wu, Xiaoshan and Yu, Yifei and Lyu, Xiaoyang and Huang, Yihua and Wang, Bo and Zhang, Baoheng and Wang, Zhongrui and Qi, Xiaojuan},
  journal={Advances in Neural Information Processing Systems},
  volume={38},
  pages={73314--73342},
  year={2026}
}

@article{weerakoon2026events,
  title={4D Reconstruction with Events \& RGB},
  author={Weerakoon, Dulanga and Rathnayake, Darshana and Do, Van Minh and Sabbella, Hemanth and Zhou, Kaichen and Zhan, Fangneng and Prakash, Alok and Sarma, Sanjay E and Misra, Archan and Liang, Paul Pu},
  journal={arXiv preprint arXiv:2610.36610},
  year={2026}
}

@inproceedings{lee2026ares,
  title={ARES: Unifying Asymmetric RGB-Event Stereo for Probabilistic Scene Flow Estimation},
  author={Lee, Jie Long and Lee, Gim Hee},
  booktitle={Proceedings of the IEEE/CVF Conference on Computer Vision and Pattern Recognition},
  pages={37022--37031},
  year={2026}
}

@inproceedings{chang2026mertracker,
  title={MER-Tracker: Towards High-Speed 3D Point Tracking via Multi-View Event-RGB Hybrid Cameras},
  author={Chang, Yiqian and Ye, Qinghong and Xu, Haoran and Li, Jianing and Ma, Dongyang and Wang, Xuan and Zhang, Wei and Tian, Yonghong and Peng, Peixi},
  booktitle={Proceedings of the IEEE/CVF Conference on Computer Vision and Pattern Recognition},
  pages={37032--37042},
  year={2026}
}

@article{guo2026interp3r,
  title={Interp3R: Continuous-time 3D Geometry Estimation with Frames and Events},
  author={Guo, Shuang and Febryanto, Filbert and Sun, Lei and Van Gool, Luc and Gallego, Guillermo},
  journal={arXiv preprint arXiv:2603.14528},
  year={2026}
}

@article{bai2026erfgs,
  title={ERF-GS: Reconstructing fast motion from disjoint event-RGB viewpoints},
  author={Bai, Xiaoyang and Li, Zhenyang and Xu, Weiwei and Lam, Edmund Y and Peng, Yifan},
  journal={Computational Visual Media},
  year={2026},
  publisher={TUP}
}

@article{fisher1921probable,
  title={On the" probable error" of a coefficient of correlation deduced from a small sample},
  author={Fisher, Ronald A},
  journal={Metron},
  volume={1},
  pages={3--32},
  year={1921}
}

\clearpage
\appendix

This appendix provides the details of the method that are omitted from the main text (\cref{sec:supp_method}), the implementation details with all parameter values (\cref{sec:supp_impl}), the details of the experimental setup (\cref{sec:supp_exp}), the complete calibration results (\cref{sec:supp_calib}), block diagrams of the three stages of the pipeline (\cref{sec:supp_diagram}), the corner detections of our Corner2Tag reimplementation (\cref{sec:supp_corner2tag}), and additional results, including dynamic 3D reconstructions and sub-frame motion for 4D slow motion (\cref{sec:supp_more}).

\section{Method Details}
\label{sec:supp_method}

\begin{figure*}[t]
    \centering
    \includegraphics[width=\linewidth]{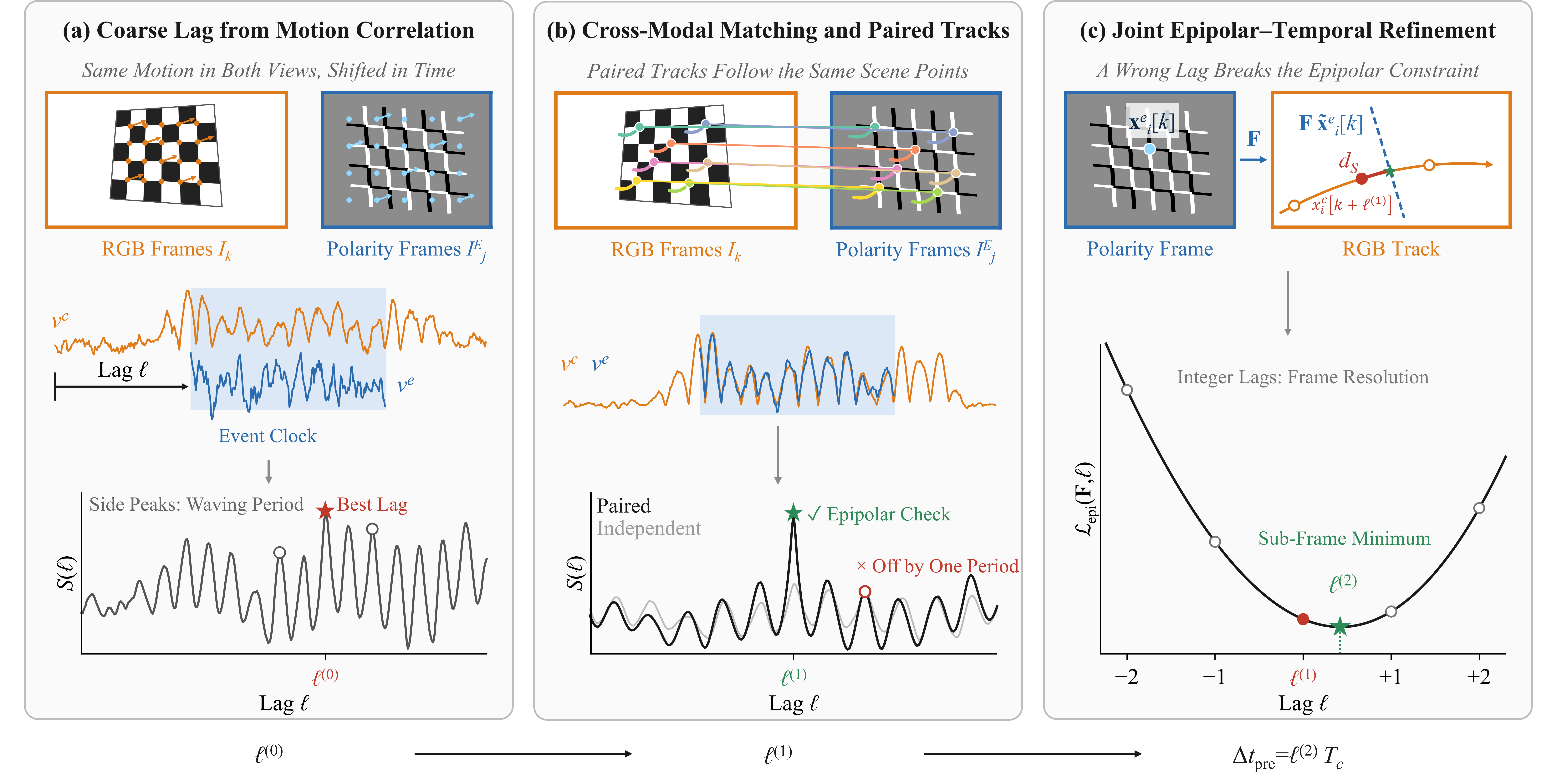}
    \caption{\textbf{Concept of the temporal pre-alignment (Sec.~3.2).}
    (a)~Independently tracked points give two motion signals that agree only at
    the correct lag, the peak of $S(\ell)$; circles mark side peaks at the waving
    period. (b)~Matched seeds are tracked in both streams, and the paired score is
    validated by the epipolar constraint. (c)~A wrong lag moves the RGB observation
    off its epipolar line; minimizing the residual over a real-valued lag gives
    $\Delta t_{\mathrm{pre}}$.}
    \label{fig:3_2illustration}
\end{figure*}

This section completes the description of the method in \cref{sec:method} and follows its order.
Symbols that are defined there are used without being introduced again. Concept illustrations of \cref{sec:supp_setup} and \cref{sec:supp_prealign} are shown in \cref{fig:3_2illustration}; \cref{sec:supp_calib_details} is shown in \cref{fig:3_3illustration}; \cref{sec:supp_depth}, \cref{sec:supp_traj} and \cref{sec:supp_subframe} are shown in \cref{fig:3_4illustration}.

\subsection{Problem Setup and Event Frames}
\label{sec:supp_setup}

\paragraph{Time model behind the constant offset.}
\Cref{eq:temporal_offset} relates the two clocks by a constant offset, which neglects the drift between the two clocks and the exposure and rolling shutter of the RGB camera.
Over the recordings that we process, which last a few seconds, the resulting timing error is bounded.
Let row $y \in [0, H)$ of RGB frame $k$ be observed at the middle of its exposure.
On the event clock, this instant is
\begin{equation}
\begin{split}
    \tau_k(y) &= (1+\drift)\,t_k - \dt_0 + \tfrac{1}{2}\Tex_k + \Trs\,\tfrac{y}{H} \\
              &= t_k - \dt + e_k(y) ,
\end{split}
\label{eq:time_model}
\end{equation}
where $\dt_0$ is the offset between the starts of the two recordings, $\drift$ the relative drift of the clocks, which includes the deviation of the frame rate from its nominal value, $\Tex_k$ the exposure time of frame $k$, and $\Trs$ the readout time of the rolling shutter.
All constant contributions are absorbed into $\dt$, and the time-varying remainder is bounded by
\begin{equation}
    |e_k(y)| \;\leq\; \tfrac{1}{2}|\drift|\,L + \tfrac{1}{2}\big(\Tex_{\max} - \Tex_{\min}\big) + \tfrac{1}{2}\,h\,\Trs ,
    \label{eq:time_bound}
\end{equation}
where $L$ is the duration of the processed recording and $h$ the fraction of the image height that the board covers.
For our recordings, $L = 9.6$\,s, so a drift of $|\drift| \leq 100$\,ppm contributes at most $0.48$\,ms.
The exposure term vanishes for a fixed exposure and is at most half the range of the exposure times under auto-exposure.
The rolling-shutter term is below $\tfrac{1}{2} h \Tc$, because the readout finishes within one frame period.
We therefore estimate only $\dt$, whose sub-frame part also absorbs the mean of $e_k(y)$, and treat the remainder as a bounded model error; if required, $\drift$ and $\Trs$ can be estimated jointly with the calibration parameters of \cref{sec:calib}.

\paragraph{Event frames.}
To apply pretrained image-domain trackers and matchers to the event stream, we render it as a sequence of \emph{event frames} by accumulating the events of short time windows into images.
We use two renderings, one before and one after the calibration.
Before the offset is known (\cref{sec:prealign}), we use \emph{polarity frames} $\Epol_j$: on a gray background, each pixel is set to white or black according to the polarity of the last event that it received in the window.
Here $j$ indexes the native event-frame sequence: $\Epol_j$ accumulates events in $[j\Te, j\Te+\Wpol)$ on the event clock.
The windows have length $\Wpol$, and a new window starts every $\Te < \Wpol$, so consecutive windows overlap.
A polarity frame is time-stamped with the start of its window, so its content is centered $\Wpol/2$ later; this constant shift is not corrected here and is absorbed by the residual offset $\res$ of \cref{sec:calib}.
The period $\Te$ is not tied to the RGB frame period $\Tc$; where the two streams have to be compared, we resample the event tracks in time (\cref{sec:supp_prealign}).
The same sequence $\{\Epol_j\}$ is reused for independent motion tracking, cross-modal matching, paired tracking, and re-seeding; refining the offset changes the frame associations, not the rendering.
After the calibration (\cref{sec:fourd}), we use \emph{count frames}: each stores the number of events per pixel, regardless of polarity, within a short window centered at a target time on the RGB clock using $\dtfin$.
We write $\Ecnt_k$ for a count frame centered at RGB time $t_k$, and $\Ecnt_{k,s}$ for one at the sub-frame time $t_{k,s}$ defined in \cref{sec:supp_subframe}; their window widths are specified in \cref{tab:params,tab:params_fourd}.
Because the calibrated offset places the window exactly at the target time, and because a count image does not invert its appearance when the motion reverses, count frames are easier to track.
Event frames are used only where a pretrained network needs an image.

\subsection{Temporal Pre-Alignment}
\label{sec:supp_prealign}

\paragraph{Query points and motion signals.}
We run CoTracker3~\cite{karaev2025cotracker3} independently on the RGB frames $\{I_k\}$ and polarity frames $\{\Epol_j\}$.
The query points are chosen separately for each modality and are not assumed to correspond, since the coarse alignment compares only the amount of motion in the two views.
In the RGB video, the queries are the checkerboard corners detected in the middle frame among the frames with a detection.
In $\{\Epol_j\}$, where no corner can be detected, a regular $G \times G$ grid covers the image region in which events occur; because a track on event frames rarely survives more than a few seconds, the grid is placed at four frames spread evenly over the recording.
All queries are tracked forward and backward in time, and a point counts as visible in a frame if the visibility predicted by the tracker exceeds $0.5$.
Since $\Te$ differs from $\Tc$ in general, we resample the event tracks to the RGB frame period by linear interpolation in time; a resampled point counts as visible only if it is visible at both neighboring samples.
Only the tracks are resampled; $j$ still indexes native polarity frames, while $k$ indexes the common-period track samples used below.
Let $\xc_i[k]$ denote the position of RGB track $i$ in frame $k$, and $\xe_i[k]$ the resampled position of event track $i$ at the event time $k\,\Tc$, so that both are sampled with period $\Tc$.
We summarize the motion in each view by the mean displacement of its visible points,
\begin{equation}
    v^{c}[k] = \frac{1}{|\mathcal{V}^{c}_k|} \sum_{i \in \mathcal{V}^{c}_k} \big\| \xc_i[k+1] - \xc_i[k] \big\|_2 ,
    \label{eq:mean_speed}
\end{equation}
where $\mathcal{V}^{c}_k$ contains the tracks visible at both samples $k$ and $k+1$; $v^{e}[k]$ is defined analogously.
A sample without any visible track is marked as unmeasured and excluded from all statistics below, instead of being counted as zero motion.
Both signals are driven by the same scene motion, mainly that of the hand-held board, and differ by the unknown time shift and, approximately, by a view-dependent gain.

\paragraph{Correlation score.}
For an integer lag $\lag$, measured in RGB frames, we compare event sample $k$ with RGB sample $k + \lag$, which by \cref{eq:temporal_offset} corresponds to the hypothesis $\dt = \lag\,\Tc$.
Let $\mathcal{I}_{\lag}$ be the set of indices $k$ for which both samples are measured, and $N_{\lag} = |\mathcal{I}_{\lag}|$.
The normalized cross-correlation of the two signals at this lag is
\begin{equation}
    \mathrm{NCC}(\lag) = \frac{1}{N_{\lag}} \sum_{k \in \mathcal{I}_{\lag}}
    \frac{v^{c}[k+\lag] - \mu^{c}_{\lag}}{\sigma^{c}_{\lag}} \cdot
    \frac{v^{e}[k] - \mu^{e}_{\lag}}{\sigma^{e}_{\lag}} ,
    \label{eq:motion_ncc}
\end{equation}
where $\mu^{c}_{\lag}$, $\sigma^{c}_{\lag}$ and $\mu^{e}_{\lag}$, $\sigma^{e}_{\lag}$ are the mean and standard deviation of the two signals over the overlap $\mathcal{I}_{\lag}$.
Standardizing over the overlap makes the correlation invariant to the gain and bias between the signals, and hence to the different resolutions and focal lengths of the two cameras.
The size of the overlap, however, changes with the lag, and a short overlap can produce a high correlation by chance.
This is why we score a lag by the Fisher $z$-statistic of its correlation, $\Score(\lag)$ in \cref{eq:score}, which is approximately standard normal for uncorrelated signals and thus measures how significant a correlation is instead of how large it is.
Since the samples of a motion signal are correlated in time, $\Score$ overstates the significance in absolute terms, and we use it only to rank lags.
The coarse lag is $\lag^{(0)} = \argmax_{\lag \in \Lambda} \Score(\lag)$, where $\Lambda$ contains the lags with
$\ell\,T_c \in [-5\,\mathrm{s},\, L_c - L_e + 1\,\mathrm{s}]$ and
$N_\ell \ge \max(8,\, n/4)$; $L_c$ and $L_e$ are the durations of the RGB and the event
recording, and $n$ is the number of measured samples of the shorter signal.

\begin{figure}[t]
    \centering
    \includegraphics[width=\columnwidth]{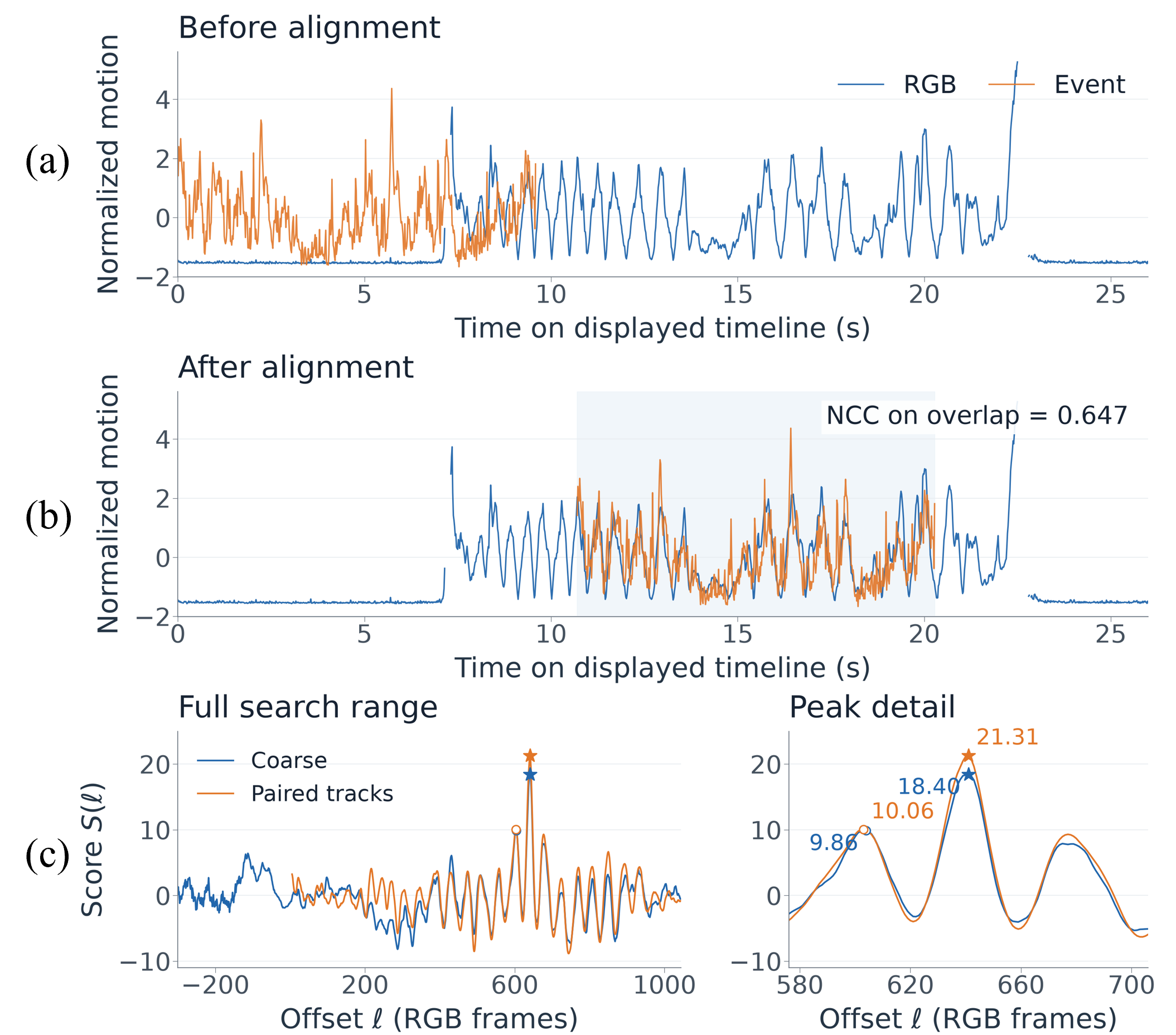}
    \caption{\textbf{Motion-based temporal alignment} on a real recording with a nominal baseline of 10\,cm.
        (A)~RGB and event motion signals, extracted independently, before alignment.
        (B)~The same signals after shifting the event signal by the coarse lag $\lag^{(0)} = 641$ RGB frames (10.694\,s).
        The shaded region marks the temporal overlap, on which $\mathrm{NCC} = 0.647$.
        Both panels use the same per-modality normalization.
        (C)~Score $\Score(\lag)$ of \cref{eq:score} for the independent tracks (coarse) and for the paired tracks, over all candidate lags $\lag$ (left) and around the dominant peak (right).
        Stars and open circles mark the highest and the second-highest distinct local peak.
        The side peaks are spaced by the waving period of the board, about 37 frames.
        The paired tracks widen the margin between the correct lag and the best side peak, from 18.4 against 9.9 to 21.3 against 10.1.}
    \label{fig:motion_alignment}
\end{figure}

\paragraph{Periodic motion.}
A hand-held board is typically waved back and forth, which makes the motion signals roughly periodic and produces side peaks of $\Score$ at multiples of the waving period (\cref{fig:motion_alignment}).
The motion is never strictly periodic, however: its amplitude, pauses, and tempo vary, and only the correct lag aligns all of these variations.
In the example of \cref{fig:motion_alignment}, the highest side peak reaches about half the score of the correct lag.
We therefore move the board irregularly, and we validate the selected lag geometrically with the paired tracks introduced next; strictly periodic motion would remain ambiguous for any motion-based alignment.

\paragraph{Cross-modal matching: frame pairs, filtering, and ranking.}
Given $\lag^{(0)}$, we pair $\Epol_j$ with the RGB frame $I_k$ whose timestamp is nearest $j\Te + \lag^{(0)}\Tc$, accounting for the different frame periods.
To narrow this gap before matching, we render each candidate RGB frame in two ways that resemble a polarity frame: as a crop around the detected board, which removes the static background that triggers no events, and as an event-like image of the signed log-intensity difference between its two neighboring frames.
We then detect and describe keypoints with SuperPoint~\cite{detone2018superpoint} and match them with LightGlue~\cite{lindenberger2023lightglue} between $\Epol_j$ and the rendered RGB frame. Each rendering is matched separately and yields its own candidate, with the event-like
image rendered in both signs; the candidates of all renderings are ranked together.
Because the board is planar, correct matches on it are related by a homography; for each frame pair, we fit one with RANSAC, keep a spatially well-distributed subset of its inliers, and rank the frame pairs by the number, the inlier ratio, and the spatial spread of these matches.
In detail, we sample one polarity frame every $0.1$\,s over the overlap given by $\lag^{(0)}$ and match it against the RGB frames at the lags $\lag^{(0)} + \{-6, -3, 0, 3, 6\}$, which gives about 480 frame pairs for a recording of 9.6\,s.
The board crop is enlarged by 25\% and resized to at most 800\,px, and the homography of a frame pair is fitted with RANSAC at a threshold of 3\,px.
Among its inliers, we keep matches in the order of decreasing LightGlue score and skip every match that lies within 4\,px of an already kept one in the event image, until at most 120 matches are kept.
A frame pair with $n_{\mathrm{s}}$ kept matches and a homography inlier ratio $r_{\mathrm{H}}$ is ranked by
\begin{equation}
    R = n_{\mathrm{s}}\, r_{\mathrm{H}}\,
    \Big(\tfrac{1}{2} + \tfrac{1}{2}\sqrt{\min(1, 4 s_{\mathrm{A}})}\Big)
    \Big(\tfrac{1}{2} + \tfrac{1}{2}\sqrt{s_{\mathrm{I}}}\Big)
    \Big(\tfrac{1}{2} + \tfrac{1}{2}\, s_{\mathrm{O}}\Big) ,
    \label{eq:match_rank}
\end{equation}
where $s_{\mathrm{A}}$ is the area of the convex hull of the kept matches relative to the image area, $s_{\mathrm{I}}$ the ratio of the smaller to the larger eigenvalue of their covariance, and $s_{\mathrm{O}}$ the fraction of occupied cells of a $4 \times 4$ grid; each of the three is computed in both images, and the smaller value is used.
The three best-ranked frame pairs that are at least $0.5$\,s apart are validated in turn, as described next.
If the highest peak of $\Score$ yields no frame pair with at least 8 kept matches, the matching is repeated at the second-highest peak. If none of the validated frame pairs is accepted, the alignment is reported as failed.

\paragraph{Paired tracks and validation of the lag.}
The best-ranked frame pairs are validated in turn.
The matches of a frame pair seed $M$ query pairs, which CoTracker3 tracks forward and backward through $\{I_k\}$ and the same $\{\Epol_j\}$ from their respective query frames.
This yields \emph{paired tracks} $\{(\xc_i[k], \xe_i[k])\}_{i=1}^{M}$, in which a common index now denotes the same scene point.
Evaluating \cref{eq:mean_speed,eq:motion_ncc,eq:score} on the paired tracks, where both signals measure the same points, gives the lag $\lag^{(1)}$, still at frame resolution.
This second search is again global over $\Lambda$: $\lag^{(0)}$ only selects the frames that are matched and does not constrain $\lag^{(1)}$, and if the two disagree, we keep $\lag^{(1)}$.
We accept the frame pair if a fundamental matrix fitted with RANSAC to its paired tracks
at the lag $\ell^{(1)}$ has an inlier ratio of at least $0.2$ and if at least four pairs
are visible in both views in at least $5\%$ of the overlapping frames.
At a lag that is off by one period of the motion, the two tracks of a pair observe the board in different poses and violate the epipolar constraint, so this test also guards against the side peaks of $\Score$.

\paragraph{Joint epipolar--temporal refinement.}
The fundamental matrix $\Fund$ maps a point of the event image to its epipolar line in the RGB image.
For a moving point, an error in the lag displaces the RGB observation along its image trajectory and violates the epipolar constraint in proportion to the velocity component normal to the epipolar line.
We evaluate the RGB tracks at non-integer indices by linear interpolation, $\xc_i(k+\lag) = (1-\alpha)\,\xc_i[\lfloor k+\lag \rfloor] + \alpha\,\xc_i[\lfloor k+\lag \rfloor + 1]$ with $\alpha = k + \lag - \lfloor k+\lag \rfloor$.
The complete objective adds a coverage term to the mean epipolar residual $\Lepi$ of \cref{eq:joint_refine}:
\begin{equation}
    \min_{\Fund,\, \lag} \;\; \Lepi(\Fund, \lag) + \lambda_{\mathrm{cov}} \max\big(0,\; \eta - |\Omega_{\lag}| / |\Omega|\big) ,
    \label{eq:joint_refine_full}
\end{equation}
where the squared Sampson distance $d_{\mathrm{S}}$~\cite{hartley2003multiple} in $\Lepi$ is measured in px$^2$, $\Omega$ is the set of all track--sample pairs, and $\Omega_{\lag} \subseteq \Omega$ contains those visible in both views at lag $\lag$.
The hinge term prevents the residual from being reduced trivially by shrinking the temporal overlap below a fraction $\eta$.
We initialize $\Fund$ with RANSAC on all paired points at lag $\lag^{(1)}$ and minimize \cref{eq:joint_refine_full} over $\Fund$ and $\lag$ with Adam; after every step, $\Fund$ is rescaled to unit Frobenius norm and projected onto the rank-two matrices.
The minimizer $\lag^{(2)} \in \mathbb{R}$ gives the pre-aligned offset $\dtpre = \lag^{(2)}\,\Tc$.
The distances are computed on the raw pixel coordinates of the tracks, because the distortion of the event camera is still unknown at this stage.
$\Fund$ is therefore an approximation of the epipolar geometry: it serves for the sub-frame pre-alignment and for gating correspondences, whereas the final offset is estimated from raw events with the full camera model in \cref{sec:calib}.

Two properties make this problem well-posed.
A single view of the planar board is degenerate for $\Fund$, but all frames share the same $\Fund$, and the correspondences accumulated while the board changes depth and orientation are not coplanar.
Moreover, the lag is constrained only by motion across the epipolar lines, so the board should be moved in more than one direction.

\paragraph{Per-frame correspondences.}
The calibration of \cref{sec:calib} is initialized from cross-modal correspondences in as many frames as possible, whereas paired tracks end when points become occluded or leave the field of view.
Whenever fewer than $M_{\min}$ pairs remain visible, we therefore re-match $I_k$ with the polarity frame $\Epol_j$ minimizing $|j\Te+\dtpre-t_k|$, keep the matches within $\theta_{\mathrm{epi}}$ pixels of their epipolar lines under $\Fund$, and track them as new pairs through the same RGB and polarity sequences.
For every RGB frame $k$, this yields a set $\Corr_k = \{(\xc_{k,h}, \xe_{k,h})\}_h$ of pairs that are visible in frame $k$ and in its closest polarity frame: $\xc_{k,h}$ is the position of pair $h$ in $I_k$, and $\xe_{k,h}$ is its position in that polarity frame.
These pairs are not interpolated in time.
The closest polarity frame differs from the RGB frame by up to $\Te/2$, in addition to the stamp shift of $\Wpol/2$ (\cref{sec:supp_setup}), which displaces a point on the moving board by a few pixels.
This is sufficient for their purpose: the pairs only initialize the event camera and anchor the first stage of the calibration, both of which require an accuracy of half a board square, and the precise timing is left to the residual offset $\res$, which is estimated from raw events.

\subsection{Raw-Event Calibration}
\label{sec:supp_calib_details}

\begin{figure*}[t]
    \centering
    \includegraphics[width=\linewidth]{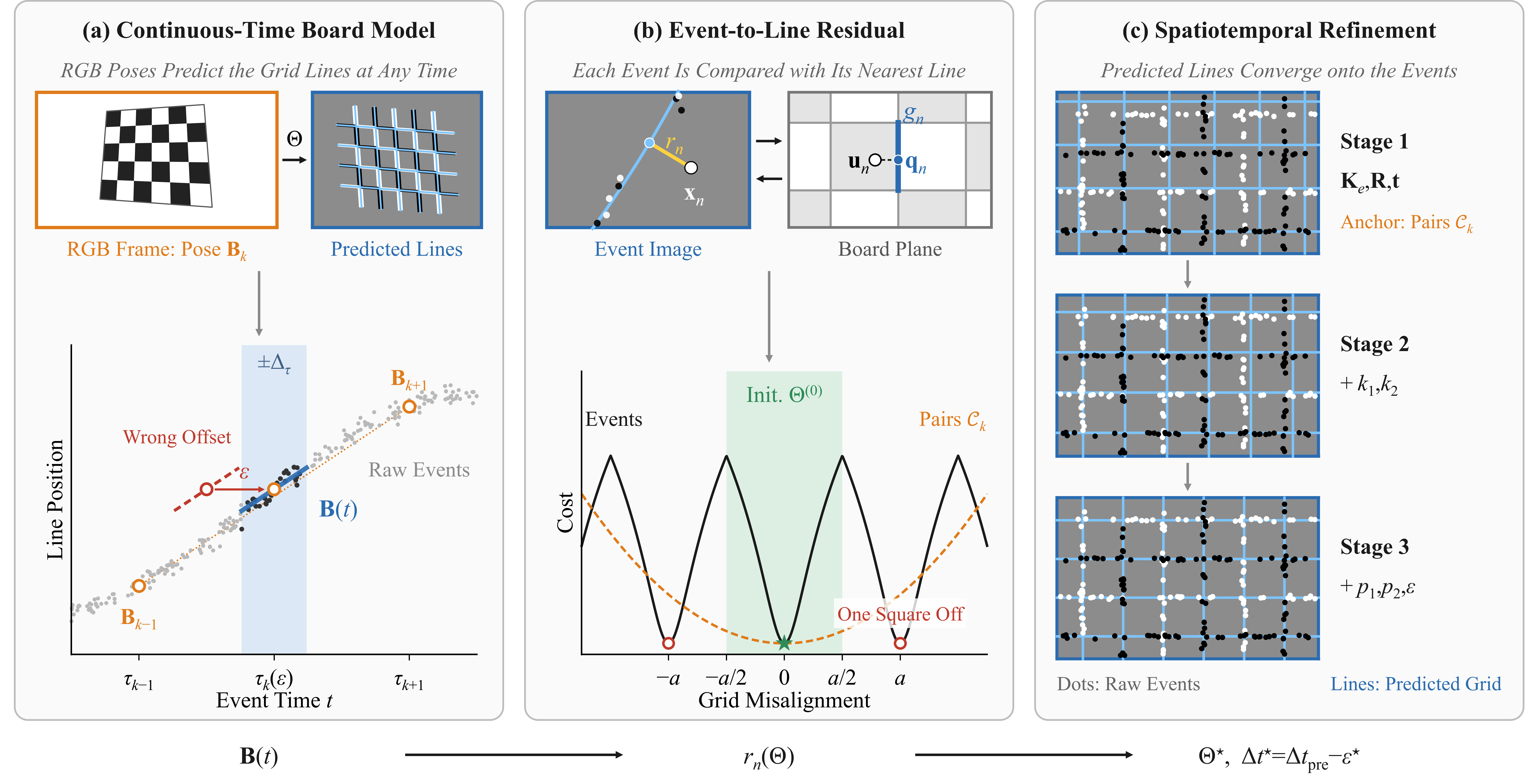}
    \caption{\textbf{Concept of the raw-event calibration (Sec.~3.3).}
    (a)~RGB board poses, extended by a first-order motion model, predict each grid
    line at any event time; a wrong residual offset $\varepsilon$ shifts the
    prediction off the events. (b)~Each event is back-projected to the board,
    assigned to its nearest grid line, and scored by its image distance $r_n$;
    the periodic grid requires an initialization within half a square.
    (c)~Over three stages, the predicted lines converge onto the events.}
    \label{fig:3_3illustration}
\end{figure*}

\paragraph{Continuous-time board model.}
The unknowns $\Par = \{\Ke, \de, \Rce, \tce, \res\}$ comprise four intrinsic, four distortion, six pose, and one temporal parameter, while $\Kc$ and $\dc$ stay fixed.
We attach a coordinate frame to the board such that the pattern lies in the plane $Z = 0$, and denote by $\mathbf{u} = (u, v)^{\top}$ the metric in-plane coordinates of a board point and by $\tilde{\mathbf{u}} = (u, v, 1)^{\top}$ their homogeneous form.
For each RGB frame $k$ in which the corners are detected, planar PnP gives the board rotation, with columns $\mathbf{r}^{1}_k, \mathbf{r}^{2}_k, \mathbf{r}^{3}_k$, and the board origin $\mathbf{o}_k$ in RGB-camera coordinates.
Since board points have zero height, the third column does not enter the $3 \times 3$ matrix $\Bd_k = [\,\mathbf{r}^{1}_k \;\; \mathbf{r}^{2}_k \;\; \mathbf{o}_k\,]$.
We estimate the board velocity by the central difference of the two adjacent frames, $\dot{\Bd}_k = (\Bd_{k+1} - \Bd_{k-1}) / (2\Tc)$; if only one neighbor has a valid pose, we use the one-sided difference, and frames without a valid neighbor are skipped.
The residual offset $\res$ in the capture time $\tau_k(\res) = t_k - \dtpre + \res$ absorbs the sub-frame error that remains in $\dtpre$, including the stamp shift of the polarity frames (\cref{sec:supp_setup}).
The half-width $\twin$ of the motion model in \cref{eq:board_motion} is a few milliseconds, much shorter than the frame period.
The model extrapolates the rotation columns linearly instead of on $\mathrm{SO}(3)$.
Within $\twin$, a hand-held board rotates by a fraction of a degree, and the deviation from the exact rotation is of second order in this angle, i.e., far below a pixel.
In \cref{eq:event_projection}, $\Hb(t)\,\tilde{\mathbf{u}}$ are the event-camera coordinates of the board point, and $\Hb(t)$ is the time-varying homography from the board plane to the normalized image plane of the event camera.
The residual offset enters \cref{eq:event_projection} through $\tau_k(\res)$.

\paragraph{Coarse initialization: steps, selection, and rejection rules.}
Registering events to a periodic grid requires an initial guess that is accurate to within half a square, which we obtain from the correspondences $\Corr_k$.
For each pair whose RGB point lies on the board, we back-project $\xc_{k,h}$ with $\Kc$ and $\dc$, intersect its viewing ray with the board plane given by $\Bd_k$, and obtain its board coordinates $\mathbf{u}_{k,h}$.
The pairs $(\mathbf{u}_{k,h}, \xe_{k,h})$ are planar 3D--2D correspondences for the event camera, as if it had observed known target points.
Zhang's closed-form solution~\cite{zhang2000flexible}, without distortion, then gives an initial $\Ke$ and the board pose in the event camera for every frame.
Composing each of these poses with the inverse RGB board pose of the same frame gives per-frame estimates of $(\Rce, \tce)$, which we fuse robustly by averaging the rotations after outlier removal and taking the median translation.
A robust reprojection refinement of $\Ke$, $\Rce$, and $\tce$ over all correspondences completes the initial estimate $\Par^{(0)}$.
A matched RGB point is considered to lie on the board if it back-projects into the printed pattern, extended by half a square, and if the pair is an inlier of a per-frame homography from the board plane to the event image (RANSAC, 4\,px).
A frame is used if it has at least 6 such pairs, and at least 3 frames are required.
Before the rotations are averaged, we discard the frames whose rotation deviates from the median rotation by more than $3^\circ$ or three times the median deviation, whichever is larger.
The final refinement uses a Cauchy loss with a scale of 2\,px.

\paragraph{Event selection.}
For each RGB frame $k$ used for calibration, we select the raw events $\mathcal{E}_k = \{n : |t_n - \tau_k(\res)| \leq \twin\}$.
To reject events that are not caused by the board, such as those of the hand that holds it, we keep only the events inside the intersection of a quadrilateral fitted to the densest event cluster and the projected board outline; we also remove hot pixels and keep at most $N_{\max}$ events per frame.
To find the board region of a frame, we accumulate the events of a 10\,ms window into an image, remove hot and isolated pixels, dilate the image so that the board forms a single blob, fit a quadrilateral to the largest blob, and intersect it with the board outline projected with the current parameters.
A pixel is considered hot if its event count over the whole recording exceeds 20 times the median count.
Frames with fewer than 150 selected events are skipped, and frames with more than $N_{\max}$ events are subsampled uniformly at random.
The sets $\mathcal{E}_k$ are selected at the beginning of each optimization stage, with the current value of $\res$, and are kept fixed within the stage.

\paragraph{Association rules.}
A moving checkerboard triggers events along its interior grid lines $\Grid$, where black and white squares meet, i.e., along the lines $u = i\,a$ and $v = j\,a$ in board coordinates.
We back-project each event onto the board plane at its own timestamp,
\begin{equation}
    \tilde{\mathbf{u}}_n \sim \Hb(t_n)^{-1}\, \proje^{-1}(\mathbf{x}_n) ,
    \label{eq:event_backprojection}
\end{equation}
where $\proje^{-1}$ returns the undistorted normalized homogeneous coordinates of a pixel, and associate it with the nearest grid line $g_n \in \Grid$.
Let $\mathbf{q}_n$ be the foot point of $\mathbf{u}_n$ on $g_n$.
Events that fall outside the pattern or close to an X-junction, where two crossing lines make the association ambiguous, are discarded: an event is dropped if it back-projects outside the printed pattern or if it lies within $0.08\,a$ of a junction along both board axes.
The outer border of the pattern is not used: events near it lie about one square from the nearest interior line and are removed by the rejection threshold below.
During a solve, only the association of an event, i.e., its line $g_n$ and its foot point $\mathbf{q}_n$ on the board, is fixed; the normal $\mathbf{n}_n$ in the residual of \cref{eq:line_distance} is recomputed from the current projection of the line at every evaluation of the residual.
The residual does not use the polarity: an edge triggers ON or OFF events depending on its direction of motion, but in both cases the events lie on the same line.
Because the grid is periodic, an association is correct only if the current prediction error is below half a square, which the coarse initialization ensures.
As in ICP~\cite{besl1992method}, we alternate: the associations $(g_n, \mathbf{q}_n)$ are fixed while $\Par$ is optimized and are recomputed between solves, and events with $|r_n|$ above a threshold $\theta_r$ are rejected.

\paragraph{Spatiotemporal refinement: schedule of the stages.}
Each stage minimizes \cref{eq:calib_objective} with a trust-region least-squares solver.
The first stage optimizes $\mathbf{K}_e$, $\mathbf{R}$, and $\mathbf{t}$ and keeps
$\varepsilon$ at its initial value; its weight is $\beta_1 = \beta\, N_E / (2 N_C)$, where
$N_E$ and $N_C$ are the numbers of associated events and of correspondences, so that the
correspondence term weighs $\beta$ times the event term irrespective of their sizes.
The second stage adds the radial coefficients $(k_1, k_2)$, and the third adds the tangential coefficients $(p_1, p_2)$ and the residual offset, constrained to $|\res| \leq \Tc$; both use $\beta_2 = \beta_3 = 0$, i.e., the correspondences anchor only the first stage.
The loss scale $\sigma_s$ and the rejection threshold $\theta_r$ decrease from stage to stage, and each stage runs a few ICP rounds.
The distortion is initialized at zero and $\res$ at $+\Tc/2$.
The final recording offset combines both temporal stages, $\dtfin = \dtpre - \res^{\star}$, where $\res^{\star}$ is the estimate of the last stage.

\subsection{Event-Guided Depth}
\label{sec:supp_depth}

\begin{figure*}[t]
    \centering
    \includegraphics[width=\linewidth]{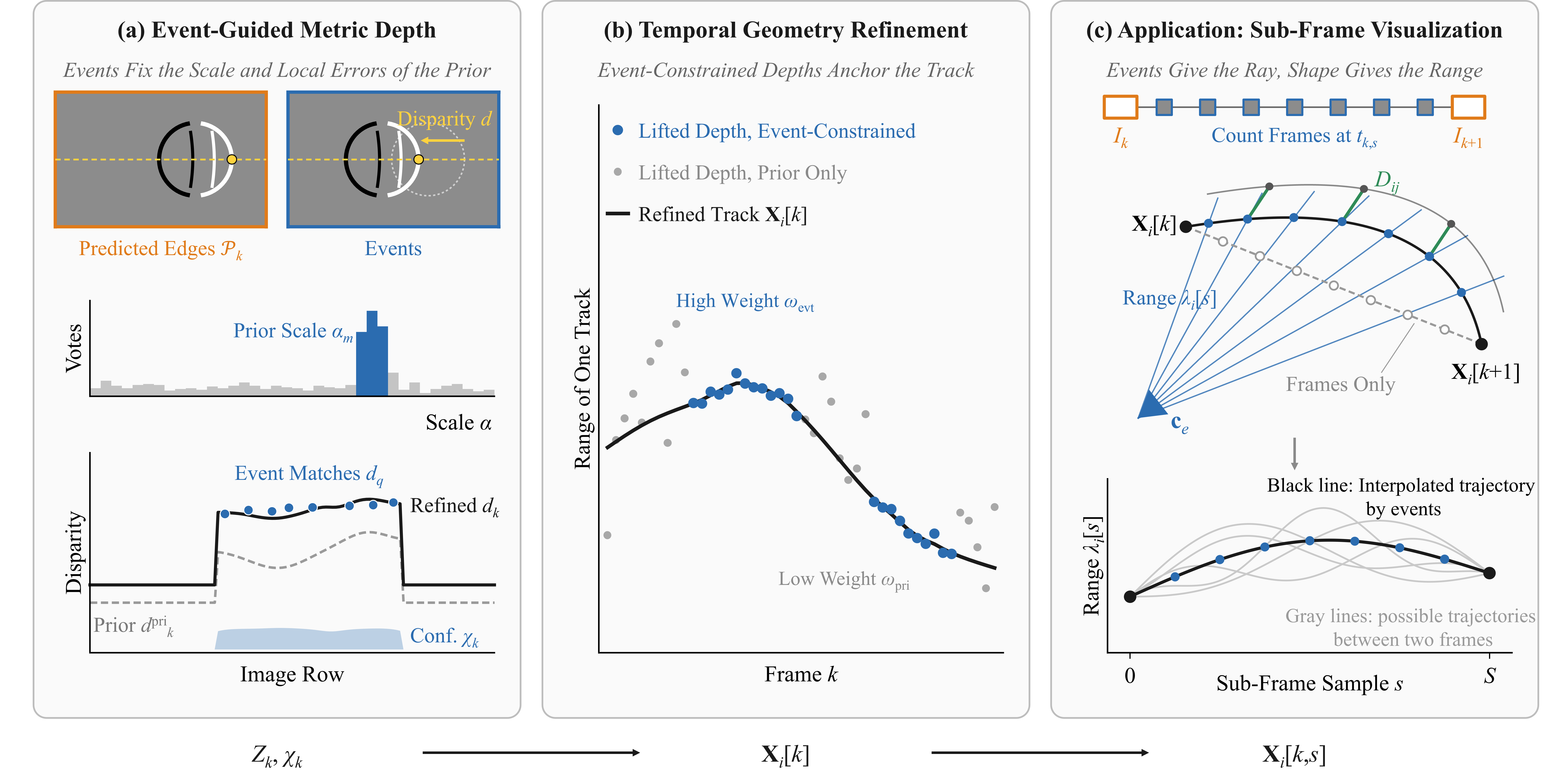}
    \caption{\textbf{Concept of the event-constrained dynamic 3D geometry (Sec.~3.4).}
    (a)~Same-row, same-polarity pairs of predicted edges and events vote for the
    prior scale $\alpha_m$; event matches then correct the disparity locally.
    (b)~Per-frame lifted depths are weighted by their source, so that the refined
    track follows the event-constrained ones. (c)~Between two RGB frames, event
    tracks fix the ray of each point, and shape preservation determines its range.}
    \label{fig:3_4illustration}
\end{figure*}

The calibrated rig is a metric stereo pair with a common time axis, using $\dtfin$ to map events to the RGB clock.
Its views remain heterogeneous: events encode brightness changes only where the scene moves, so dense intensity-based stereo matching does not apply.
We instead use them as sparse metric constraints on a dense RGB depth prior, first correcting its scale and then its local errors.

\paragraph{Prior and predicted event edges.}
The rig remains static while it observes the dynamic scene.
The depth prior $Z^{\mathrm{pri}}_k$ is predicted by a pretrained 3D foundation model.
Its metric scale is approximate, and its errors vary locally.
We rectify the calibrated pair such that a pixel $\mathbf{x}$ of the rectified RGB image with depth $Z$ appears in the rectified event image on the same row, displaced by the disparity $d = f b / Z$.
All corrections are applied to $d$, i.e., in inverse depth, starting from the prior disparity $d^{\mathrm{pri}}_k = f b / Z^{\mathrm{pri}}_k$.
Let $L_k$ be the log intensity of frame $k$ and $\mathbf{w}_k$ its optical flow to the next frame, estimated with RAFT~\cite{teed2020raft}.
Over a window of half-width $\dwin$ around the frame time, the linearized event generation model~\cite{gallego2020event} predicts the brightness change
\begin{equation}
    \Delta L_k(\mathbf{x}) \approx - \frac{2\dwin}{\Tc}\, \nabla L_k(\mathbf{x})^{\top} \mathbf{w}_k(\mathbf{x}) .
    \label{eq:predicted_events}
\end{equation}
The edge pixels of frame $k$ with $|\Delta L_k(\mathbf{x})| > \theta_L$ form the predicted event edges $\mathcal{P}_k$, each with the expected polarity $\sign \Delta L_k(\mathbf{x})$.
We take the logarithm of the gamma-encoded intensity; since gamma encoding is a power law, this only rescales $\theta_L$ and changes neither the location nor the polarity of the predicted edges.
The events with $|t_n - t_k| \leq \dwin$ must coincide with $\mathcal{P}_k$ displaced by the disparity, which we enforce at two levels.

\paragraph{Depth scale: voting and consistency across chunks.}
First, we estimate one global scale $\alpha_m$ of the prior disparity for each temporal chunk $\mathcal{T}_m$ of the video by epipolar voting: every event, paired with every predicted edge pixel of the same polarity on its rectified row, votes for the scale $\alpha = d / d^{\mathrm{pri}}_k(\mathbf{x})$ that their disparity $d$ implies, and the scale is taken from the peaks of the resulting histogram.
The chunks are the inference chunks of the depth prior, whose scales can differ, and we choose the scales of all chunks jointly such that they agree at the chunk boundaries.
The votes for the scale are collected in a histogram over $\log \alpha$; 
a pair of an event and a predicted edge pixel votes only if its disparity corresponds to a depth between $0.5$ and $10$\,m
and each event votes at most once per bin.
We take up to four peaks of the smoothed histogram as candidates, fit a few frames of the chunk with each of them, and record the support $\nu_m(\alpha)$ of a candidate, i.e., the number of events that it associates, normalized by that of the best candidate of the chunk.
A chunk comprises 24 frames and overlaps its successor by 8 frames.
At each boundary, we measure the ratio $r_m$ between the prior disparities of chunks $m$ and $m+1$ on the moving pixels, so that consistent scales satisfy $\alpha_{m+1} = \alpha_m r_m$, and select one candidate per chunk by dynamic programming,
\begin{equation}
    \min_{\{\alpha_m\}} \; \sum_{m} - \log \nu_m(\alpha_m) + \sum_{m} \frac{\big( \log \alpha_{m+1} - \log \alpha_m - \log r_m \big)^{2}}{2 \sigma_{\alpha}^{2}} ,
    \label{eq:scale_dp}
\end{equation}
with $\sigma_{\alpha} = 0.07$.

\paragraph{Local depth correction: matching, regions, and kernels.}
Second, we correct the scaled prior locally.
Starting from the disparity that the scaled prior predicts, we match every event to the nearest predicted edge pixel of the same polarity on its row, within regions bounded by depth and motion discontinuities, and refine the matches in a few ICP-style rounds with a decreasing gate.
Each match $q$ provides a measured disparity $d_q$ at a pixel $\mathbf{x}_q$ and thus the relative error $c_q$ of the scaled prior that enters \cref{eq:depth_update}.
The weight in \cref{eq:depth_update,eq:confidence} is $\kappa_q(\mathbf{x}) = w_q\, \kappa_{\mathrm{s}}\big(\|\mathbf{x} - \mathbf{x}_q\|\big)\, \kappa_{\mathrm{d}}(\mathbf{x}, \mathbf{x}_q)$; it combines a spatial kernel $\kappa_{\mathrm{s}}$, a kernel $\kappa_{\mathrm{d}}$ that compares the prior depths of $\mathbf{x}$ and $\mathbf{x}_q$, and the reliability $w_q$ of the match.
A match therefore influences only nearby pixels of similar prior depth, so corrections do not leak across depth discontinuities.
The prior weight $w_0$ makes the correction fade to the scaled prior away from matches.
The event confidence $\chi_k(\mathbf{x})$ in \cref{eq:confidence} measures the relative influence of the events in this correction and supplies the source-aware depth weights of \cref{sec:supp_traj}.
We set $w_q$ to the squared horizontal component of the unit normal of the matched edge, because a nearly horizontal edge constrains the horizontal disparity only weakly.
A match is searched within $\max(6\,\mathrm{px}, 18\%)$ of the disparity predicted by the scaled prior, and the three rounds use gates of 4, 2.5, and 1.5\,px.
The regions are the connected components, of at least 400\,px, that are bounded by jumps of the prior above 6\% and by flow discontinuities; they separate static from moving surfaces, and within each region an affine mapping of the prior is fitted to the event matches.
The spatial kernel $\kappa_{\mathrm{s}}$ is a Gaussian with a standard deviation of 12\,px, and $\kappa_{\mathrm{d}}(\mathbf{x}, \mathbf{x}_q) = 1$ if the prior inverse depths of $\mathbf{x}$ and $\mathbf{x}_q$ fall into the same or into adjacent bins of 5\% width, and $0$ otherwise.
The dense correction of \cref{eq:depth_update} does not use the region boundaries, because $\kappa_{\mathrm{d}}$ already keeps corrections from crossing depth discontinuities.

\paragraph{Photometric disparity on fast-moving surfaces.}
On fast-moving surfaces, the RGB edges are motion-blurred and $\mathcal{P}_k$ becomes unreliable.
There we estimate the disparity photometrically.
Events that are warped into the RGB view with the correct disparity explain the blur, so the event-based double integral (EDI)~\cite{pan2019bringing} recovers a sharp image.
We apply this test on tiles of $32 \times 32$ pixels, with a stride of 16, that are mostly moving, contain events, and have a blur length $\|\mathbf{w}_k\|\, \Tex / \Tc$ of at least 3\,px, where $\Tex$ is the exposure time.
For each such tile $\Gamma$, we search a small range around the current estimate $d_0$ for the disparity that gives the sharpest reconstruction,
\begin{equation}
    d_{\Gamma} = \argmin_{|d - d_0| \,\leq\, \Delta_d} \; \frac{\big\| \nabla \hat{I}^{(d)}_k \big\|_{1,\Gamma}}{\big\| \nabla \hat{I}^{(d)}_k \big\|_{2,\Gamma}} ,
    \label{eq:edi_search}
\end{equation}
where $\hat{I}^{(d)}_k$ is the EDI reconstruction of frame $k$ from the events displaced by $d$.
The ratio of the $L_1$ to the $L_2$ norm of its gradients over the tile is small for a sharp image and does not depend on the image contrast.
Each $d_\Gamma$ replaces the edge matches inside its tile and enters \cref{eq:depth_update} in their place.
EDI needs the exposure time of the RGB camera and the contrast threshold $C$ of the event camera, and we estimate both from the data: $\Tex$ maximizes the correlation between the signed event count in an exposure window centered at the frame time and the brightness change predicted from the flow, and $C$ is the slope of this fit.

\paragraph{Static background and sources of depth.}
Finally, since the rig is static, the static background triggers no events and cannot be corrected by them.
We identify it as those regions of the local correction whose median flow magnitude is below $0.4$\,px per frame, and set its disparity to the temporal median of the per-frame estimates.
The median replaces the current estimate only where the two differ by less than 25\%, so that an object that pauses keeps its own depth.
Every pixel thus has one of three sources of depth: the static background, a moving surface that is constrained by events, with $\chi_k(\mathbf{x}) \geq \tfrac{1}{2}$, or a moving surface that relies on the prior.

\subsection{Persistent 3D Trajectories}
\label{sec:supp_traj}

\paragraph{Query grid and occlusion boundaries.}
We track a dense grid of points through the RGB video with CoTracker3, adding new queries where tracks are lost, and lift each visible track point with the depth of its frame, discarding the points on occlusion boundaries.
The query grid has a spacing of $1/128$ of the image width, as in Stereo4D~\cite{jin2025stereo4d}.
Every 10 frames, new queries are added in the grid cells that contain no visible track.
A pixel is considered to lie on an occlusion boundary if the range of the depth in its $3 \times 3$ neighborhood exceeds 30\% of its depth.

\paragraph{The two terms adopted from Stereo4D.}
In \cref{eq:fourd_objective}, $\Lsta^{i}$ penalizes the spread of the positions of track $i$ over time, $\Lacc^{i}$ penalizes its acceleration along the ray, and $\mu_i \in [0,1]$, a soft indicator of a static track computed from its 2D motion, balances them.
Both terms and $\mu_i$ follow Eqs.~(1)--(3) of Stereo4D~\cite{jin2025stereo4d}.
The acceleration is evaluated with time steps of 1, 3, and 5 frames, and $\mu_i$ decreases as a sigmoid of the 90th percentile of the largest 2D displacement of the track within 16 frames, with the threshold of Stereo4D, 20\,px at an image width of 512\,px, scaled to our resolution.
We deviate in two details: each term is normalized by the mean squared range of the track and by its number of summands, and the acceleration is evaluated only where all three frames are observed.

\paragraph{Source-aware depth term.}
Stereo4D regularizes all offsets equally toward the measured depth, whereas our depth comes from sources of different reliability.
As in Stereo4D, the regularizer acts on the inverse range, but we weight it by the source of each lifted point,
\begin{equation}
    \Ldep = \sum_{i} \sum_{k} \omega_i[k] \left( \frac{1}{\|\mathbf{X}_i[k]\|} - \frac{1}{\|\mathbf{X}^{0}_i[k]\|} \right)^{2} ,
    \label{eq:depth_term}
\end{equation}
\begin{equation}
    \omega_i[k] =
    \begin{cases}
        \omega_{\mathrm{bg}} & \text{on the static background,} \\[2pt]
        \omega_{\mathrm{pri}}^{\,1-\chi}\; \omega_{\mathrm{evt}}^{\,\chi} & \text{otherwise, with } \chi = \chi_k(\mathbf{x}_i[k]) ,
    \end{cases}
    \label{eq:depth_weight}
\end{equation}
with $\omega_{\mathrm{evt}} > \omega_{\mathrm{bg}} > \omega_{\mathrm{pri}}$.
The measured depth is thus trusted most where it is constrained by events and least where it relies on the prior alone, and the geometric interpolation by the event confidence avoids a hard switch between the two.

\paragraph{Shape term and its track pairs.}
We segment the moving objects with SAM~3~\cite{carion2026sam}, prompted automatically by the moving tracks, and penalize changes in the 3D distance $D_{ij}[k] = \|\mathbf{X}_i[k] - \mathbf{X}_j[k]\|$ of track pairs on the same object,
\begin{equation}
    \Lshp = \sum_{(i,j) \in \mathcal{N}} \sum_{k} \rho_{\sigma}\big( D_{ij}[k] - \bar{D}_{ij} \big) ,
    \label{eq:rigid}
\end{equation}
where $\mathcal{N}$ is the set of such pairs and $\bar{D}_{ij} = \med_{k} D_{ij}[k]$ is the median distance of a pair over time.
The median is recomputed from the current positions at every iteration, so the term penalizes only changes of a distance over time and not the distance itself.
The Cauchy loss tolerates deviations from rigidity, such as articulated or deforming objects.
The set $\mathcal{N}$ is built every 10 frames: each track on an object is paired with its 6 nearest neighbors in 3D and with 4 randomly chosen tracks of the same object, and a pair is kept if both tracks are visible together in at least 5 frames.
The random pairs link distant parts of an object, which nearest neighbors alone would leave unconstrained with respect to each other.

\paragraph{Event-view term and its partner tracks.}
Each moving RGB track $i$ receives a partner track $\mathbf{y}_i[k]$ in the event camera: we project the lifted point into the event image at one frame and track it with CoTracker3 through count frames $\{\Ecnt_k\}$ (\cref{sec:supp_setup}) centered at the RGB frame times.
A partner track is initialized at the frame in which the 2D speed of its RGB track is closest to 6\,px per frame, among the frames with a valid depth whose lifted point projects into the event image.
At this speed, the point moves fast enough to trigger events and slowly enough for a sharp RGB frame and a reliable depth.
Let $h_i[k]$ be the distance between $\mathbf{X}_i[k]$ and the viewing ray of $\mathbf{y}_i[k]$, and $\bar{h}_i$ its mean over the track.
We penalize its variation over time,
\begin{equation}
    \Levt = \sum_{i} \sum_{k \in \mathcal{K}_i} \rho_{\sigma}\!\left( \frac{h_i[k] - \bar{h}_i}{b} \right) ,
    \label{eq:cross}
\end{equation}
where $\mathcal{K}_i$ contains the frames in which the partner lies within 2\,px of the epipolar line of $\mathbf{x}_i[k]$, and $\bar{h}_i$ is eliminated in closed form at every iteration.
We ask for a constant rather than a zero distance because the partner track is initialized with the current, still inaccurate depth and may therefore follow a neighboring surface point; a constant distance remains consistent with a correct trajectory, whereas depth errors that vary over time change $h_i[k]$.
The constancy is an approximation: if the partner follows a point at a distance $\Delta$ from the tracked one, a rotation of the surface by an angle $\theta$ changes $h_i[k]$ by about $\Delta\,\theta$.
The epipolar gate and the Cauchy loss limit the influence of such observations.
We minimize \cref{eq:fourd_objective} jointly over all offsets with Gauss--Newton iterations.

\subsection{Sub-Frame Trajectories}
\label{sec:supp_subframe}

The optimized trajectories are sampled at RGB frame times, while events constrain the motion between them (\cref{fig:pipeline}, top right).
As a downstream application of the refined geometry, the following steps construct event tracks through both endpoint projections and recover the remaining range along each event ray; the resulting sub-frame trajectories are rendered as novel-view 4D slow motion, the application mentioned in \cref{sec:fourd}.

\paragraph{Count frames and event tracks between two RGB frames.}
Between the RGB frames $k$ and $k+1$, we render count frames $\Ecnt_{k,s}$ at the sub-frame times $t_{k,s} = t_k + (s/S)\,\Tc$, $s = 0, \dots, S$.
The upsampling factor $S$ is adaptive: it is chosen for each interval such that the moving points advance by about one pixel per sample, up to $S = 64$.
We project every moving point at both end frames into the event image, $\mathbf{y}_i[k,0] = \proje(\Rce\,\mathbf{X}_i[k] + \tce)$ and $\mathbf{y}_i[k,S] = \proje(\Rce\,\mathbf{X}_i[k+1] + \tce)$, and track it with CoTracker3 through the count frames $\{\Ecnt_{k,s}\}_{s=0}^{S}$, forward from the first projection and backward from the second.
We blend the two tracks linearly, with the weights $1 - s/S$ and $s/S$, so that the resulting event track $\mathbf{y}_i[k,s]$ passes through both projections.
The window of $\Ecnt_{k,s}$ spans twice the spacing of the samples, clipped to the range of 1 to 10\,ms, so the windows of consecutive samples overlap by about half.
A point has a valid event track if its forward and its backward track each end within 2\,px of the opposite projection.

\paragraph{Range along the event rays.}
Each $\mathbf{y}_i[k,s]$ defines a viewing ray of the event camera, with origin at the event-camera center $\mathbf{c}_e = -\Rce^{\top}\tce$ and unit direction $\mathbf{g}_i[k,s]$ in RGB-camera coordinates, so that
\begin{equation}
    \mathbf{X}_i[k,s] = \mathbf{c}_e + \lambda_i[s]\, \mathbf{g}_i[k,s] .
    \label{eq:subframe_ray}
\end{equation}
The events thus determine the rays, i.e., they measure the motion orthogonal to the viewing direction directly, whereas the range $\lambda_i[s]$ cannot be observed from a single view.
We resolve it with the same shape prior as above: each point should keep its distances to the neighboring points on the same object.
We recover the unknown ranges by minimizing
\begin{multline}
    \Lsub = \sum_{(i,j) \in \mathcal{N}} \sum_{s=1}^{S-1} \rho_{\sigma}\big( \|\mathbf{X}_i[k,s] - \mathbf{X}_j[k,s]\| - D_{ij}[k,s] \big) \\
    + \gamma \sum_{i} \sum_{s=1}^{S-1} \ddot{\lambda}_i[s]^{2} ,
\label{eq:subframe_objective}
\end{multline}
with the end ranges $\lambda_i[0]$ and $\lambda_i[S]$ fixed by $\mathbf{X}_i[k]$ and $\mathbf{X}_i[k+1]$.
Here $\rho_{\sigma}$ is the Cauchy loss, $D_{ij}[k,s]$ interpolates linearly between the distances $D_{ij}[k]$ and $D_{ij}[k+1]$ measured at the two RGB frames, and $\ddot{\lambda}_i[s] = S^{2} \big( \lambda_i[s-1] - 2\lambda_i[s] + \lambda_i[s+1] \big)$ is the acceleration along the ray with time measured in frame periods, so that the weight $\gamma$ does not depend on $S$.
The weight is small, $\gamma = 10^{-4}$ for lengths in meters: an acceleration of 0.1\,m per squared frame period costs as much as a distance error of 1\,mm, so the second term only decides what the shape term leaves open.

\paragraph{Points without a valid event track.}
A point without a valid event track takes the deviation from the straight segment between $\mathbf{X}_i[k]$ and $\mathbf{X}_i[k+1]$ of up to 8 validly tracked neighbors on the same object within 5\,cm, averaged with inverse-distance weights; without such neighbors, it follows the straight segment.
Each point keeps the color of the RGB frame at which it was queried, and rendering the points at the instants $t_{k,s}$ from novel viewpoints yields 4D slow motion.

\section{Implementation Details and Parameters}
\label{sec:supp_impl}
\begin{table*}[t]
\centering
\caption{\textbf{Parameters of \method:} event frames, temporal pre-alignment, and calibration.}
\label{tab:params}
\footnotesize
\setlength{\tabcolsep}{5pt}
\begin{tabular}{@{}llll@{}}
\toprule
Stage & Parameter & Symbol & Value \\
\midrule
Event frames & Window of a polarity frame & $\Wpol$ & 15\,ms \\
 & Period of the polarity frames & $\Te$ & $1/90$\,s \\
 & Window of a count frame (centered) & -- & 10\,ms at the RGB frame times; see \cref{tab:params_fourd} between frames \\
\midrule
Pre-alignment & RGB queries & -- & board corners of the middle frame with a detection \\
(\cref{sec:prealign}) & Event query grid; frames & $G$ & $10 \times 10$; frames $0$, $N/4$, $N/2$, $3N/4$ \\
 & Visibility threshold & -- & 0.5 \\
 & Lag search range; minimum overlap & $\Lambda$ & $\ell\,T_c \in [-5\,\mathrm{s},\; L_c - L_e + 1\,\mathrm{s}]$; $N_\ell \ge \max(8,\, n/4)$ \\
 & Frame pairs for matching & -- & every 0.1\,s; lags $\lag^{(0)} + \{-6,-3,0,3,6\}$ \\
 & Board crop & -- & enlarged by 25\%, at most 800\,px \\
 & Homography RANSAC threshold & -- & 3\,px \\
 & Match spacing; maximum; minimum number & -- & 4\,px; 120; 8 \\
 & Validated frame pairs; temporal spacing & -- & 3; at least 0.5\,s \\
 & Acceptance: inlier ratio of $\mathbf{F}$; co-visible pairs & -- &
  $\ge 0.2$; $\ge 4$ in $\ge 5\%$ of the frames \\
 & RANSAC threshold for the initial $\Fund$ & -- & 3\,px \\
 & Adam iterations; schedule; learning rate & -- & 800; cosine; $10^{-2}$ for $\ell$, $10^{-6}$ for $\mathbf{F}$ \\
 & Coverage weight; minimum coverage & $\lambda_{\mathrm{cov}}$, $\eta$ & 10; 0.15 \\
 & Re-seeding threshold; epipolar gate & $M_{\min}$, $\theta_{\mathrm{epi}}$ & 4; 3\,px \\
\midrule
Calibration & PnP reprojection RMS for a valid frame & -- & $\leq 1.5$\,px \\
(\cref{sec:calib}) & Frame stride; maximum; validation split & -- & 2; 160; every fifth frame \\
 & On-board margin; homography RANSAC & -- & half a square; 4\,px \\
 & Minimum pairs per frame; minimum frames & -- & 6; 3 \\
 & Rotation outlier threshold & -- & $3^\circ$ or $3\times$ the median deviation, whichever is larger \\
 & Cauchy scale of the initial refinement & -- & 2\,px \\
 & Half-width of the raw-event window & $\twin$ & 2\,ms \\
 & Window for the board region; hot-pixel factor & -- & 10\,ms; 20 times the median count \\
 & Events per frame: minimum; maximum & --, $N_{\max}$ & 150; 1500 \\
 & Junction exclusion radius & -- & $0.08\,a$ \\
 & Cauchy scales of the three stages & $\sigma_s$ & 3, 1.5, 1\,px \\
 & Rejection thresholds of the three stages & $\theta_r$ & 10, 5, 3\,px \\
 & ICP rounds of the three stages & -- & 4, 3, 3 \\
 & Weight of the correspondence term & $\beta$ & 0.3 (first stage only) \\
 & Residual offset: initial value; bound & $\res$ & $+\Tc/2$; $|\res| \leq \Tc$ \\
\bottomrule
\end{tabular}
\end{table*}

\begin{table*}[t]
\centering
\caption{\textbf{Parameters of \method, continued:} depth and 4D reconstruction (\cref{sec:fourd}).}
\label{tab:params_fourd}
\footnotesize
\setlength{\tabcolsep}{5pt}
\begin{tabular}{@{}llll@{}}
\toprule
Stage & Parameter & Symbol & Value \\
\midrule
Depth & Half-width of the event window & $\dwin$ & 2.5\,ms \\
 & Threshold on the brightness change & $\theta_L$ & 0.12 \\
 & Chunk length; overlap & -- & 24 frames; 8 frames \\
 & Scale candidates per chunk; smoothness & --, $\sigma_{\alpha}$ & up to 4; 0.07 \\
 & Search range of a match & -- & $\max(6\,\mathrm{px}, 18\%)$ of the predicted disparity \\
 & Gates of the three rounds & -- & 4, 2.5, 1.5\,px \\
 & Regions: prior jump; minimum size & -- & 6\%; 400\,px \\
 & Spatial kernel; depth bins & $\kappa_{\mathrm{s}}$, $\kappa_{\mathrm{d}}$ & Gaussian, 12\,px; 5\% \\
 & Prior weight & $w_0$ & 0.5 \\
 & EDI tiles: size; stride; minimum blur & $\Gamma$ & $32 \times 32$\,px; 16\,px; 3\,px \\
 & EDI search range & $\Delta_d$ & 6\,px \\
 & Exposure time; contrast threshold & $\Tex$, $C$ & estimated from the data, once per recording \\
 & Static background: median flow of a region & -- & below 0.4\,px per frame \\
 & Background median accepted & -- & within 25\% of the current estimate \\
 & Event-constrained pixel & $\chi_k$ & $\geq 0.5$ \\
\midrule
Trajectories & Query grid spacing; re-query interval & -- & $1/128$ of the image width; 10 frames \\
 & Occlusion boundary & -- & depth range in the $3 \times 3$ neighborhood above 30\% \\
 & Time steps of the acceleration term & -- & 1, 3, 5 frames \\
 & Static indicator & $\mu_i$ & sigmoid of the 90th percentile of the largest 2D displacement \\
 & & & within 16 frames; threshold 20\,px at a width of 512\,px \\
 & Source weights & $\omega_{\mathrm{evt}}, \omega_{\mathrm{bg}}, \omega_{\mathrm{pri}}$ & 20, 1, 0.1; $\omega_\mathrm{evt} = 5$ for tracks on a segmented object \\
 & Term weights & $\lambda_{\mathrm{dep}}, \lambda_{\mathrm{shp}}, \lambda_{\mathrm{evt}}$ & 1, 20, 20 \\
 & Track pairs: nearest; random; interval; co-visible & $\mathcal{N}$ & 6; 4; 10 frames; at least 5 frames \\
 & Partner track: initial speed; epipolar gate & -- & closest to 6\,px per frame; 2\,px \\
 & Scales of the Cauchy losses & $\sigma$ & 2\,cm (shape); 1.5\,px of disparity (event view) \\
\midrule
Sub-frame & Upsampling factor & $S$ & about 1\,px of motion per sample, at most 64 \\
 & Window of a count frame (centered) & -- & twice the sample spacing, clipped to 1--10\,ms \\
 & Blending of forward and backward track & -- & linear, $1 - s/S$ and $s/S$ \\
 & Valid event track & -- & both tracks end within 2\,px of the opposite projection \\
 & Acceleration weight & $\gamma$ & $10^{-4}$ (lengths in m, time in frame periods) \\
 & Fallback: neighbors; radius & -- & up to 8; 5\,cm \\
\bottomrule
\end{tabular}
\end{table*}

\paragraph{Implementation details.}
All learned components are used with their public weights and are kept frozen; the parameter values are the same for all synthetic and real recordings.
For the RGB camera, we use two radial distortion coefficients and iteratively remove views with large reprojection errors.
Polarity frames use windows of $\Wpol = 15$\,ms that start every $\Te = 1/90$\,s, and count frames accumulate 10\,ms at the RGB frame times.
In the pre-alignment, the event queries form a grid with $G = 10$, placed at the frames $0$, $N/4$, $N/2$, and $3N/4$ of the $N$ polarity frames.
We estimate the initial fundamental matrix with a RANSAC threshold of 3\,px and optimize \cref{eq:joint_refine_full} for 800 Adam iterations with a cosine learning-rate schedule, $\lambda_{\mathrm{cov}} = 10$, and $\eta = 0.15$.
New pairs are seeded when fewer than $M_{\min} = 4$ pairs are visible, with an epipolar threshold of $\theta_{\mathrm{epi}} = 3$\,px.
In the calibration, we discard RGB frames with a PnP reprojection RMS above 1.5\,px and use every second remaining frame, up to 160 frames, with every fifth frame held out for validation.
The event-to-line residuals use $\twin = 2$\,ms and at most $N_{\max} = 1500$ events per frame.
The three stages use Cauchy scales $\sigma_s$ of 3, 1.5, and 1\,px, rejection thresholds $\theta_r$ of 10, 5, and 3\,px, and 4, 3, and 3 ICP rounds, with $\beta = 0.3$.
The depth module uses $\dwin = 2.5$\,ms and $\theta_L = 0.12$; its remaining parameters and those of the trajectory optimization are listed in \cref{tab:params_fourd}. All experiments are run on a single NVIDIA A100 GPU.

\Cref{tab:params,tab:params_fourd} list all parameters of \method; the same values are used for all synthetic and real recordings.

\section{Experimental Details}
\label{sec:supp_exp}
This section gives the details of the experiments that \cref{sec:experiments} summarizes.

\paragraph{Synthetic data.}
We render synthetic recordings in Blender and simulate noise-free events from the rendered frames with an ESIM-style simulator~\cite{Rebecq18corl}.
The RGB camera records $1920 \times 1080$ frames at 60\,fps, and the event camera has a resolution of $640 \times 512$.
Both cameras follow an ideal pinhole model without distortion, and a checkerboard with $7 \times 5$ inner corners and 18\,cm squares is moved in front of them at a distance of about 3\,m.
Starting from a default configuration with a 20\,cm baseline, parallel optical axes, and no temporal offset, we vary one factor at a time.
For the temporal misalignment, we shift the event recording by $\dt \in \{-0.637, +0.283, +0.912\}$\,s relative to the RGB recording, with the sign convention of \cref{eq:temporal_offset}.
We further use baselines of 5, 20, and 50\,cm and rotate the event camera toward the RGB camera by $5^\circ$, $10^\circ$, and $20^\circ$.
For the comparison with EF-Calib~\cite{wang2024ef} (\cref{tab:cocalib_syn}), we render the recordings with the offsets $\dt \in \{-0.037, +0.013, +0.044\}$\,s a second time with its circle pattern on the same board, so that both patterns follow the same trajectory.
Every recording provides the ground-truth camera parameters, temporal offset, per-frame depth of both views, and object poses.

\paragraph{Real data.}
We pair an event camera (LUCID Triton2 EVS TRT003S-EC with a Sony IMX637 sensor, $640\times512$ pixels) with a Nikon COOLPIX P1100, which records $1920\times1080$ video at 59.94\,fps with a rolling shutter and auto-exposure, at two baselines of approximately 10 and 25\,cm.
The cameras are mounted on a rigid rig without a shared trigger.
In each setting, the event camera records for 9.6\,s within an RGB video of 24--26\,s, so the recording offset amounts to several seconds.
The calibration target is a hand-held checkerboard with $9 \times 9$ inner corners and a square size of 3.1\,cm.
For the evaluation of the event-guided depth, we additionally record a hand-held checkerboard with an Intel RealSense D457 depth camera registered to the RGB camera, whose depth serves as the ground truth.

\paragraph{Baselines.}
For the intrinsic calibration of the event camera, we compare with E2Calib~\cite{muglikar2021calibrate} and Corner2Tag~\cite{ryu2026corner2tag}.
E2Calib reconstructs grayscale images from the events with E2VID~\cite{rebecq2019high} and calibrates on the reconstructed images, whereas Corner2Tag detects the checkerboard corners in images of accumulated events.
Since the code of Corner2Tag was not publicly available at the time of writing, we reimplemented it following the paper; \cref{sec:supp_corner2tag} shows the corner detections of our reimplementation.
Its results may therefore differ from those of the original implementation, and its errors in \cref{tab:calib_syn} are higher than those in its original paper; we will update them once the official code is released.
For the spatiotemporal calibration of the pair, we compare with E2Calib and EF-Calib~\cite{wang2024ef}.
E2Calib calibrates the stereo pair on the reconstructed images and assumes synchronized recordings, whereas EF-Calib jointly estimates the intrinsics, the extrinsics, and the temporal offset with a dedicated pattern recognized in the event stream.
Since E2Calib assumes synchronized recordings, we provide it with the offset $\dtpre$ from our temporal pre-alignment (\cref{sec:prealign}) on the temporally misaligned recordings and reconstruct its images at the correspondingly shifted RGB frame times.

\paragraph{Metrics on synthetic data.}
We report errors with respect to the ground truth.
For the intrinsics, these are the relative focal-length error $e_f$ (in \%, averaged over $f_x$ and $f_y$), the principal-point error $e_c = \|\hat{\mathbf{c}} - \mathbf{c}\|_2$ (in pixels), and the projection error $e_{\pi}$, i.e., the mean pixel distance between the projections of a regular grid of viewing rays that covers the field of view, with a spacing of 20\,px, under the estimated and the ground-truth camera model.
The projection error summarizes the joint effect of intrinsics and distortion, whose individual coefficients are strongly correlated.
For the extrinsics, we report the geodesic rotation error $e_{R}$ (in degrees), the translation error $e_{t} = \|\hat{\tce} - \tce\|_2$, and the baseline-length error $e_{b} = \big|\|\hat{\tce}\| - \|\tce\|\big|$ (in mm); for the temporal offset, the absolute error $e_{\dt} = |\dtfin - \dt|$ (in ms).

\paragraph{Protocol on real data.}
On real data, where the ground truth is unavailable, we repeat the calibration on 40 random subsets of each recording.
The temporal pre-alignment, the RGB intrinsics, and the board poses are computed once from the full recording; each trial then draws 24 frames at random from the frames in which all compared methods obtain a valid observation, and every method is calibrated on the same frames.
We report the mean and the standard deviation of the estimated parameters.
Since the same event camera is used in all settings, its intrinsics should agree across trials and settings.
We also report the estimated baseline $\|\hat{\tce}\|$.

\paragraph{Depth metrics.}
For depth, we report the root-mean-square error (RMSE) and the mean absolute error (MAE), both in metres, and the absolute relative error (AbsRel), i.e., the mean of $|Z - Z^{*}|/Z^{*}$, where $Z$ and $Z^{*}$ are the estimated and the ground-truth depth.
The errors are computed on the moving regions, over the pixels where the ground truth and all compared depth maps are valid, without rescaling any prediction to the ground truth.

\paragraph{Event warping in the qualitative comparison.}
In \cref{fig:warpcompare}, each event is back-projected to its viewing ray, intersected with the board plane given by the RGB pose at the time of the event, and projected into the RGB frame; only the events on the board are shown.
As E2Calib does not estimate the temporal offset, its events are warped with the offset of our temporal pre-alignment.

\paragraph{Spread of the focal length on real data.}
In \cref{tab:calib_real}, the focal length of \method varies more between trials than that of E2Calib.
This larger spread stems from the small number of frames per trial.
With 24 frames, the depth of the board varies by less than 0.25\,m, so a longer focal length and a more distant camera produce almost the same image of the board, and the focal length becomes the least constrained parameter.
Since the event-to-line residual is periodic in the grid, the objective is non-convex, and with so few frames a trial can settle in a local minimum along this direction.
The parameters that do not lie on it remain stable: the principal point varies by $1.1$--$5.3$\,px and the estimated baseline by $0.5$--$1.2$\,cm, against $8.4$--$9.8$\,px and $1.4$--$2.1$\,cm for E2Calib.
With the full recording, the board covers a wider range of depths and orientations and several times as many events constrain the solution, so the optimization reaches the correct minimum; this is the setting of \cref{tab:calib_syn}, where the focal length is within $0.13\%$ of the ground truth.

\section{Full Calibration Results}
\label{sec:supp_calib}
The main text reports the intrinsic calibration on synthetic data as errors with respect to the ground truth (\cref{tab:calib_syn}). \cref{tab:calib_syn_full} lists the estimated parameters of every case, from which these errors are computed.
For the real data, \cref{tab:calib_real_full} extends the statistics over repeated trials (\cref{tab:calib_real}) by the distortion coefficients. \Cref{tab:cocalib_raw} lists the raw estimates behind \cref{tab:cocalib_syn}.

\begin{table*}[t]
\centering
\caption{\textbf{Event-camera calibration on synthetic data: estimated parameters} (single run per case).
Misaligned: temporal offset $\dt$ between the two recordings.
Baseline: distance between the two cameras.
Rotation: rotation of the event camera toward the RGB camera.
Unless varied, the recordings use a 20\,cm baseline, parallel cameras, and no temporal offset.
Within each case, dark and light blue mark the best and the second-best result, ranked by the absolute error with respect to the ground truth; ties share a rank.}
\label{tab:calib_syn_full}
\small
\setlength{\tabcolsep}{4pt}
\resizebox{\textwidth}{!}{%
\begin{tabular}{lll cccc cccc}
\toprule
Setting & Case & Method
& \multicolumn{4}{c}{Intrinsic Parameters}
& \multicolumn{4}{c}{Distortion Coefficients} \\
\cmidrule(lr){4-7} \cmidrule(lr){8-11}
& & & $f_x$ & $f_y$ & $c_x$ & $c_y$
& $k_1$ ($\times 10^{-1}$)
& $k_2$ ($\times 10^{-1}$)
& $p_1$ ($\times 10^{-3}$)
& $p_2$ ($\times 10^{-3}$) \\
\midrule
\multicolumn{3}{l}{Ground truth}
& 620.00 & 620.00 & 320.00 & 256.00
& 0.00 & 0.00 & 0.00 & 0.00 \\
\midrule

Misaligned & $\Delta t = -0.637$\,s & E2Calib~\cite{muglikar2021calibrate}
& 618.91 & 618.85
& \secondcell{315.65} & \secondcell{256.45}
& \secondcell{$-$0.24} & \secondcell{2.10}
& \secondcell{0.79} & \secondcell{$-$2.73} \\
& & Corner2Tag~\cite{ryu2026corner2tag}
& \secondcell{619.57} & \secondcell{619.59}
& 338.26 & 273.67
& $-$0.96 & 3.11 & 7.23 & 6.85 \\
& & \method (ours)
& \bestcell{619.75} & \bestcell{619.81}
& \bestcell{319.83} & \bestcell{255.74}
& \bestcell{$-$0.09} & \bestcell{1.32}
& \bestcell{$-$0.05} & \bestcell{0.01} \\
\cmidrule(lr){2-11}

& $\Delta t = +0.283$\,s & E2Calib~\cite{muglikar2021calibrate}
& \secondcell{618.61} & \secondcell{618.61}
& \secondcell{315.39} & \secondcell{256.27}
& \secondcell{$-$0.27} & \secondcell{2.67}
& 0.82 & \secondcell{$-$2.94} \\
& & Corner2Tag~\cite{ryu2026corner2tag}
& 613.43 & 611.52 & 338.65 & 252.38
& 1.48 & $-$11.63 & \bestcell{$-$0.03} & 11.76 \\
& & \method (ours)
& \bestcell{619.41} & \bestcell{619.44}
& \bestcell{319.61} & \bestcell{256.18}
& \bestcell{$-$0.11} & \bestcell{1.50}
& \secondcell{$-$0.07} & \bestcell{0.00} \\
\cmidrule(lr){2-11}

& $\Delta t = +0.912$\,s & E2Calib~\cite{muglikar2021calibrate}
& \secondcell{619.23} & \secondcell{619.23}
& 315.93 & 256.72
& \secondcell{$-$0.34} & \secondcell{3.21}
& \secondcell{0.80} & $-$2.67 \\
& & Corner2Tag~\cite{ryu2026corner2tag}
& 618.58 & 620.90
& \secondcell{318.96} & \secondcell{255.58}
& 0.72 & $-$10.88
& $-$2.03 & \secondcell{1.05} \\
& & \method (ours)
& \bestcell{619.87} & \bestcell{619.88}
& \bestcell{319.26} & \bestcell{255.67}
& \bestcell{$-$0.04} & \bestcell{0.99}
& \bestcell{$-$0.08} & \bestcell{0.04} \\
\midrule

Baseline & 5\,cm & E2Calib~\cite{muglikar2021calibrate}
& \secondcell{618.41} & \secondcell{618.31}
& 315.99 & \secondcell{257.10}
& \secondcell{$-$0.29} & \secondcell{2.82}
& \secondcell{1.09} & \secondcell{$-$2.58} \\
& & Corner2Tag~\cite{ryu2026corner2tag}
& 632.34 & 631.53 & \secondcell{321.91} & 247.24
& 0.34 & $-$6.48 & $-$2.52 & 2.74 \\
& & \method (ours)
& \bestcell{619.48} & \bestcell{619.47}
& \bestcell{319.69} & \bestcell{255.62}
& \bestcell{0.00} & \bestcell{0.58}
& \bestcell{$-$0.02} & \bestcell{0.02} \\
\cmidrule(lr){2-11}

& 20\,cm & E2Calib~\cite{muglikar2021calibrate}
& \secondcell{618.88} & \secondcell{618.88}
& \secondcell{317.07} & \secondcell{257.52}
& \secondcell{$-$0.25} & \secondcell{1.97}
& \secondcell{1.16} & $-$2.15 \\
& & Corner2Tag~\cite{ryu2026corner2tag}
& 635.09 & 634.76 & 315.71 & 266.37
& 0.78 & $-$5.37 & 5.16 & \secondcell{$-$0.21} \\
& & \method (ours)
& \bestcell{619.17} & \bestcell{619.25}
& \bestcell{319.44} & \bestcell{255.84}
& \bestcell{0.01} & \bestcell{0.15}
& \bestcell{$-$0.09} & \bestcell{0.02} \\
\cmidrule(lr){2-11}

& 50\,cm & E2Calib~\cite{muglikar2021calibrate}
& \secondcell{617.62} & \secondcell{617.79}
& \secondcell{316.38} & \secondcell{256.79}
& $-$0.24 & \secondcell{1.16}
& 0.86 & $-$3.49 \\
& & Corner2Tag~\cite{ryu2026corner2tag}
& 614.14 & 614.36 & 325.17 & 253.77
& \secondcell{0.19} & $-$1.70
& \bestcell{$-$0.05} & \secondcell{1.86} \\
& & \method (ours)
& \bestcell{619.63} & \bestcell{619.66}
& \bestcell{319.34} & \bestcell{256.03}
& \bestcell{0.08} & \bestcell{$-$0.40}
& \secondcell{$-$0.15} & \bestcell{0.05} \\
\midrule

Rotation & $5^\circ$ & E2Calib~\cite{muglikar2021calibrate}
& \secondcell{617.62} & \secondcell{617.38}
& \secondcell{316.16} & \bestcell{257.06}
& \bestcell{0.02} & \secondcell{$-$1.22}
& \secondcell{1.01} & \secondcell{$-$2.14} \\
& & Corner2Tag~\cite{ryu2026corner2tag}
& 635.33 & 636.96 & 343.69 & 265.41
& \secondcell{2.05} & $-$26.45 & 9.60 & 16.32 \\
& & \method (ours)
& \bestcell{619.39} & \bestcell{619.40}
& \bestcell{320.15} & \secondcell{254.75}
& \bestcell{0.02} & \bestcell{0.27}
& \bestcell{0.06} & \bestcell{0.01} \\
\cmidrule(lr){2-11}

& $10^\circ$ & E2Calib~\cite{muglikar2021calibrate}
& \secondcell{619.59} & \secondcell{619.27}
& \secondcell{316.47} & \bestcell{256.83}
& \secondcell{0.08} & \secondcell{$-$0.64}
& \secondcell{0.88} & \secondcell{$-$1.88} \\
& & Corner2Tag~\cite{ryu2026corner2tag}
& 575.28 & 576.64 & 348.34 & 270.54
& 0.26 & 0.79 & 8.35 & 5.77 \\
& & \method (ours)
& \bestcell{619.84} & \bestcell{619.95}
& \bestcell{320.04} & \secondcell{254.53}
& \bestcell{$-$0.01} & \bestcell{0.16}
& \bestcell{0.11} & \bestcell{$-$0.03} \\
\cmidrule(lr){2-11}

& $20^\circ$ & E2Calib~\cite{muglikar2021calibrate}
& \secondcell{618.83} & \secondcell{618.80}
& \bestcell{318.59} & \secondcell{256.69}
& \secondcell{$-$0.14} & \secondcell{0.20}
& \secondcell{0.58} & \secondcell{1.08} \\
& & Corner2Tag~\cite{ryu2026corner2tag}
& 595.29 & 599.46 & 328.67 & 264.80
& $-$1.27 & 1.25 & 3.21 & 20.61 \\
& & \method (ours)
& \bestcell{620.91} & \bestcell{620.52}
& \secondcell{321.55} & \bestcell{255.55}
& \bestcell{0.00} & \bestcell{0.11}
& \bestcell{0.15} & \bestcell{$-$0.02} \\
\bottomrule
\end{tabular}}
\end{table*}

\begin{table*}[t]
\centering
\caption{\textbf{Event-camera calibration on real data: all estimated parameters.}}
\label{tab:calib_real_full}
\small
\setlength{\tabcolsep}{4pt}
\resizebox{\textwidth}{!}{%
\begin{tabular}{lll cccc cccc c}
\toprule
RGB Camera & Baseline & Method
& \multicolumn{4}{c}{Intrinsic Parameters}
& \multicolumn{4}{c}{Distortion Coefficients}
& Est. Baseline (cm) \\
\cmidrule(lr){4-7} \cmidrule(lr){8-11}
& & & $f_x$ & $f_y$ & $c_x$ & $c_y$
& $k_1$ ($\times 10^{-1}$)
& $k_2$ ($\times 10^{-1}$)
& $p_1$ ($\times 10^{-3}$)
& $p_2$ ($\times 10^{-3}$) & \\
\midrule
Nikon & \shortstack{$\approx10$\,cm}
& E2Calib~\cite{muglikar2021calibrate}
& $1228.60 \pm 8.41$ & $1228.28 \pm 8.71$
& $338.38 \pm 9.76$ & $249.74 \pm 6.63$
& $-2.64 \pm 0.28$ & $-2.19 \pm 2.65$
& $1.37 \pm 0.80$ & $-0.76 \pm 1.74$
& $15.47 \pm 1.36$ \\
& & \method (ours)
& $1231.47 \pm 15.38$ & $1229.94 \pm 14.66$
& $322.07 \pm 1.09$ & $252.53 \pm 2.70$
& $-3.32 \pm 2.40$ & $1.21 \pm 75.55$
& $0.24 \pm 0.24$ & $-0.18 \pm 0.10$
& $12.08 \pm 0.51$ \\
\cmidrule(lr){2-12}
& \shortstack{$\approx25$\,cm}
& E2Calib~\cite{muglikar2021calibrate}
& $1255.49 \pm 11.21$ & $1252.46 \pm 10.88$
& $337.00 \pm 8.36$ & $246.40 \pm 10.78$
& $-2.94 \pm 0.29$ & $5.10 \pm 2.66$
& $1.26 \pm 0.83$ & $0.33 \pm 2.02$
& $32.31 \pm 2.13$ \\
& & \method (ours)
& $1196.91 \pm 51.07$ & $1190.28 \pm 62.76$
& $319.63 \pm 5.26$ & $254.99 \pm 4.66$
& $-4.19 \pm 2.67$ & $0.55 \pm 79.74$
& $0.13 \pm 0.35$ & $0.10 \pm 0.15$
& $26.39 \pm 1.17$ \\
\bottomrule
\end{tabular}}
\end{table*}

\begin{table*}[t]
\centering
\caption{\textbf{Raw temporal offsets and extrinsics for synthetic Event--RGB co-calibration.}
The test cases correspond to \cref{tab:cocalib_syn}.
Ground truth, E2Calib, and \method use RGB-to-event transforms with
axis--angle rotation vectors. 
EF-Calib reports event-to-RGB transforms
with roll, pitch, and yaw angles, for which the ground-truth translation
is $(200,0,0)$\,mm. The baseline is the Euclidean norm of the translation.
}
\label{tab:cocalib_raw}
\small
\setlength{\tabcolsep}{5pt}
\resizebox{\linewidth}{!}{%
\begin{tabular}{llcccc}
\toprule
GT offset (s) & Method & Est.\ offset (s)
& Rotation ($^\circ$) & Translation (mm) & Baseline (mm) \\
\midrule
\multirow{4}{*}{$-0.037$}
& GT & $-0.037000$ & $(0,0,0)$ & $(-200.00,0,0)$ & $200.00$ \\
& E2Calib~\cite{muglikar2021calibrate}
& $-0.034749$ & $(0.067050,\,0.361096,\,-0.000079)$
& $(-200.433455,\,0.583424,\,-3.601533)$ & $200.47$ \\
& EF-Calib~\cite{wang2024ef}
& $-0.040765$ & $(-0.078,\,0.357,\,0.023)$
& $(184.133,\,-2.289,\,-8.367)$ & $184.34$ \\
& \method (ours)
& $-0.037381$ & $(0.084243,\,-0.000226,\,-0.002263)$
& $(-200.620062,\,0.829911,\,-0.563403)$ & $200.62$ \\
\midrule
\multirow{4}{*}{$+0.013$}
& GT & $+0.013000$ & $(0,0,0)$ & $(-200.00,0,0)$ & $200.00$ \\
& E2Calib~\cite{muglikar2021calibrate}
& $+0.018580$ & $(0.151766,\,0.376056,\,-0.001981)$
& $(-200.360070,\,1.669508,\,-7.318758)$ & $200.50$ \\
& EF-Calib~\cite{wang2024ef}
& $+0.013503$ & $(0.102,\,-0.242,\,-0.004)$
& $(211.869,\,3.510,\,5.722)$ & $211.98$ \\
& \method (ours)
& $+0.012614$ & $(0.049270,\,-0.011871,\,-0.003840)$
& $(-200.432612,\,0.497137,\,-1.760109)$ & $200.44$ \\
\midrule
\multirow{4}{*}{$+0.044$}
& GT & $+0.044000$ & $(0,0,0)$ & $(-200.00,0,0)$ & $200.00$ \\
& E2Calib~\cite{muglikar2021calibrate}
& $+0.051916$ & $(0.145814,\,0.314105,\,-0.005377)$
& $(-200.287898,\,2.580962,\,-1.820540)$ & $200.31$ \\
& EF-Calib~\cite{wang2024ef}
& $+0.041771$ & $(0.181,\,-0.443,\,0.001)$
& $(219.702,\,6.659,\,-5.120)$ & $219.86$ \\
& \method (ours)
& $+0.043588$ & $(0.063676,\,-0.014580,\,-0.003566)$
& $(-200.576336,\,0.139390,\,-1.188835)$ & $200.58$ \\
\bottomrule
\end{tabular}%
}
\end{table*}

\section{Block Diagrams of the Pipeline}
\label{sec:supp_diagram}
\Cref{fig:pipeline} gives an overview of \method; this section adds one concept diagram for each of its three stages.
Fig.~\ref{fig:3_2illustration}, ~\ref{fig:3_3illustration} and ~\ref{fig:3_4illustration} introduce concept illustrations of \cref{sec:prealign}, \cref{sec:calib} and \cref{sec:fourd} respectively.
Each diagram has three panels that follow the steps of its stage in the order of \cref{sec:method}, from the inputs of the stage to its output. \cref{sec:supp_method} gives the full description of every step.

\section{Corner2Tag Reimplementation}
\label{sec:supp_corner2tag}
\begin{figure*}[t]
    \centering
    \includegraphics[width=0.8\linewidth]{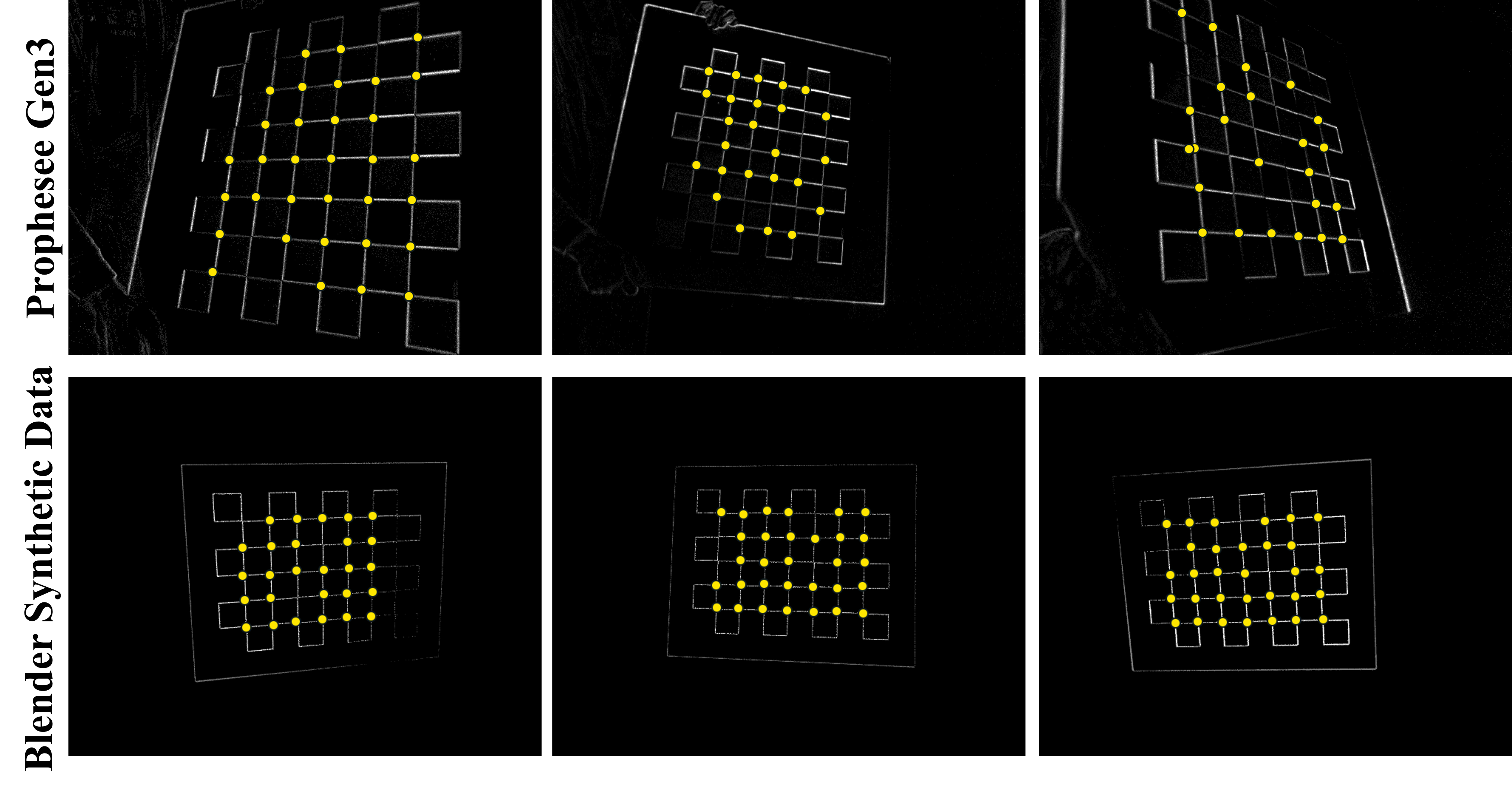}
    \caption{\textbf{Corner2Tag reproduction results.} From Prophesee\cite{muglikar2021calibrate} and one of our synthetic scene.}
    \label{fig:Corner2TagReproduction}
\end{figure*}

\paragraph{Reimplementation of Corner2Tag.}
As no official implementation of Corner2Tag was publicly available at the time of writing, we reimplemented the detector following the description in Ryu et al.~\cite{ryu2026corner2tag}. A paper cannot specify every implementation detail, so for the steps below we adopted the following settings. (1) \emph{Motion compensation}: events are warped to the midpoint of each time window, and the motion parameters are estimated by maximizing the variance of the unsigned image of warped events (IWE), using a coarse grid search followed by Powell optimization. (2) \emph{Candidate corner patches}: the polarity pattern is evaluated on concentric rings with radii of 3, 6, and 9 pixels, and a candidate is kept if the response is consistent at two or more radii. (3) \emph{Corner initialization}: seeds are selected as local intensity maxima within a two-pixel-wide band along each patch boundary; when the propagation yields four clusters at more than one step, the last such step is used to fit the two lines whose intersection gives the initial corner. These settings are our own and may differ from those of the original implementation, so all Corner2Tag results in this paper should be read as results of our reimplementation rather than of the authors' original code. Fig.~\ref{fig:Corner2TagReproduction} shows the corners detected by our reimplementation on a synthetic checkerboard sequence and on the E2Calib Prophesee Gen3 sequence.

\section{Additional Results}
\label{sec:supp_more}
\begin{figure*}[t]
    \centering
    \includegraphics[width=\linewidth]{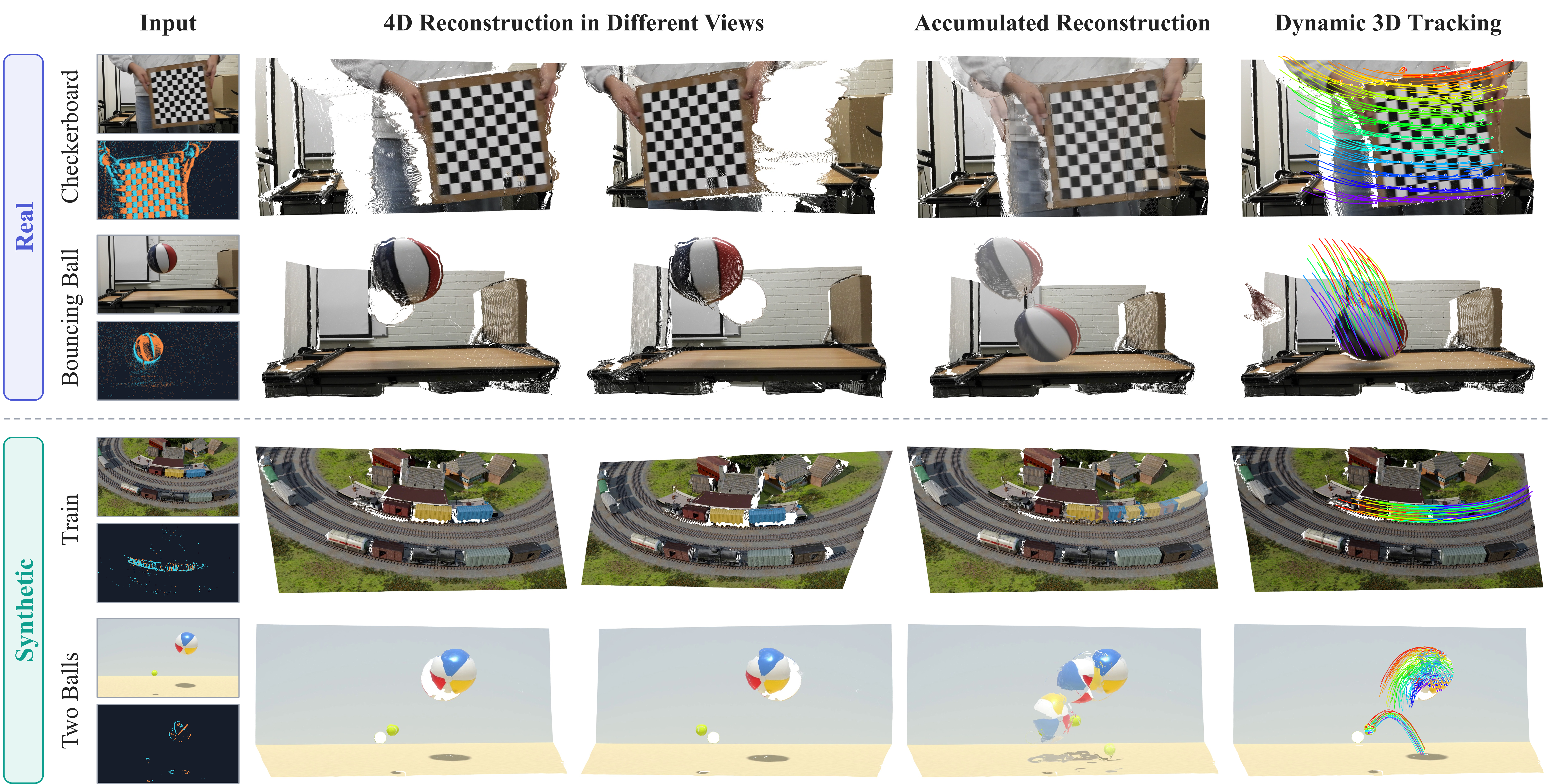}
    \caption{\textbf{Dynamic 3D reconstruction on real and synthetic scenes.} For each scene, from left to right: an RGB frame and the events around it, the reconstructed geometry of one instant rendered from two novel viewpoints, the reconstructions of several instants accumulated in one view, and the 3D trajectories of tracked points.}
    \label{fig:dynamic_recon}
\end{figure*}

\begin{figure*}[t]
    \centering
    \includegraphics[width=\linewidth]{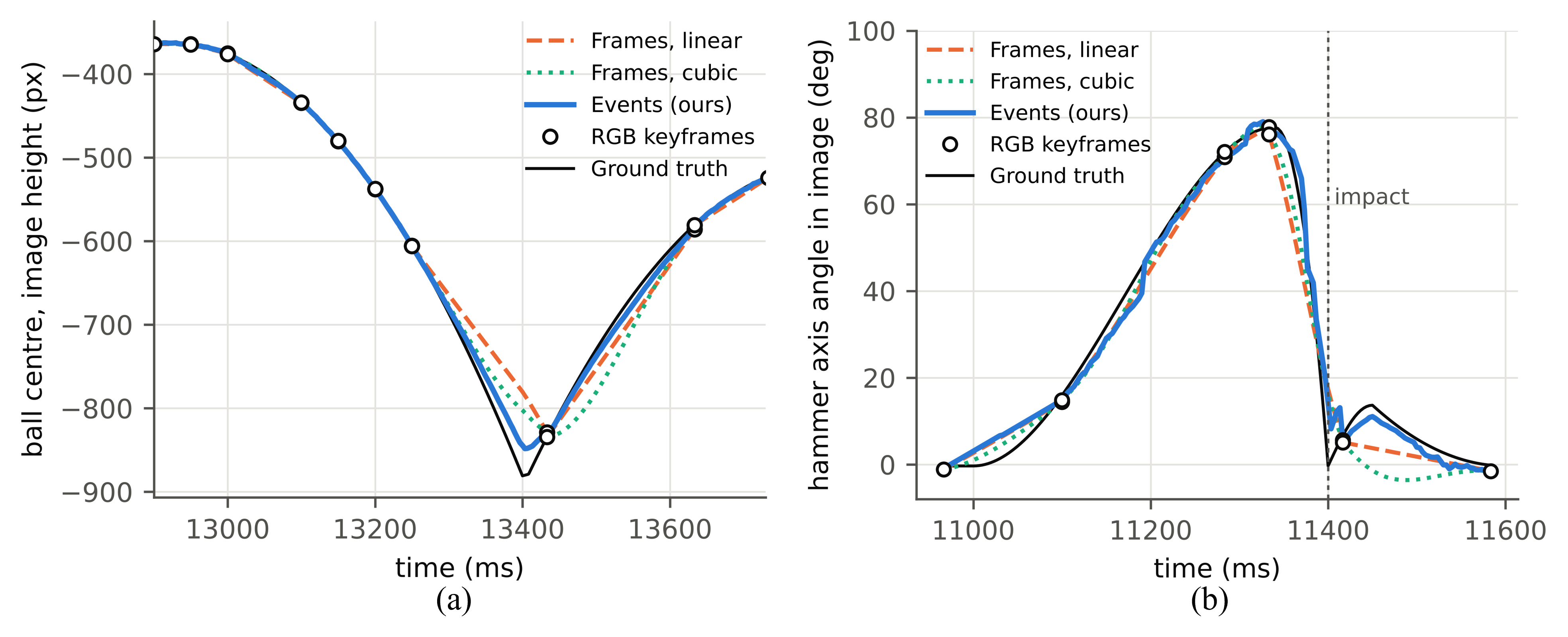}
    \caption{\textbf{Motion between keyframes on the synthetic scenes.} Circles mark the RGB keyframes; all curves pass through the same keyframe reconstructions and differ only in between. (a)~Image height of the beach-ball centre while the ball bounces between two keyframes that are 11 frames (183\,ms) apart. (b)~Image-plane orientation of the hammer while it is raised, strikes (60\,ms), and rebounds; the impact falls between two keyframes that are 5 frames (83\,ms) apart. Linear and cubic interpolation of the keyframes cut through the bounce and start the strike too early, and cubic interpolation overshoots after the impact. The event-driven trajectory (ours) follows the ground truth.}
    \label{fig:subframe_traj}
\end{figure*}

\begin{figure*}[t]
    \centering
    \includegraphics[width=0.8\linewidth]{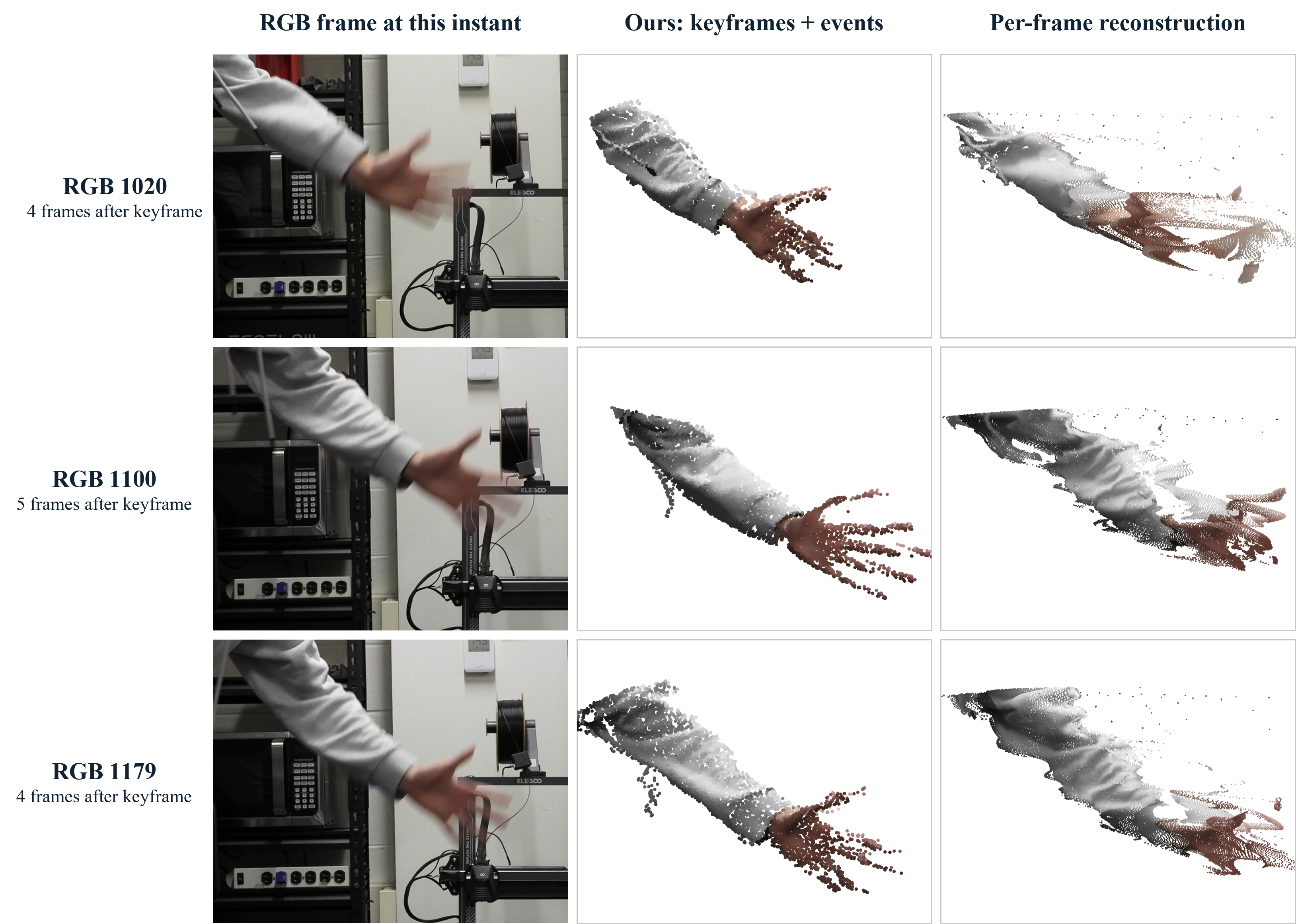}
    \caption{\textbf{Sub-frame reconstruction of fast motion on real data.} A fast-moving arm is blurred in the RGB frames between the keyframes in which it is sharp (left). We reconstruct the geometry at the sharp keyframes and carry it to the instants in between with the events (middle), and compare with the reconstruction obtained from the blurred frame itself (right).}
    \label{fig:realworldslowmotioncompare}
\end{figure*}

This section shows the results that \cref{sec:exp_fourd} refers to: dynamic 3D reconstructions on real and synthetic scenes (\cref{fig:dynamic_recon}) and the sub-frame motion that underlies the 4D slow motion (\cref{fig:subframe_traj,fig:realworldslowmotioncompare}).
The supplementary package also contains the corresponding videos.

\end{document}